%% file: main.tex
\documentclass[sn-basic]{sn-jnl}

\usepackage{graphicx}%
\usepackage{multirow}%
\usepackage{amsmath, amssymb, amsfonts}%
\usepackage{amsthm}%
\usepackage{mathrsfs}%
\usepackage[title]{appendix}%
\usepackage{textcomp}%
\usepackage{manyfoot}%
\usepackage{booktabs}%
\usepackage{algorithm}%
\usepackage{algorithmicx}%
\usepackage{algpseudocode}%
\usepackage{listings}%
\usepackage{comment}
\usepackage{xcolor}

\usepackage{mathtools}
\usepackage{dsfont}
\usepackage{pgfplots}

\pgfplotsset{compat=1.16} 
\newtheorem{theorem}{Theorem}
\newtheorem{proposition}[theorem]{Proposition}%

\input{definitions}

\begin{document}

\title[Model-agnostic variable importance for predictive uncertainty]{Model-agnostic variable importance for predictive uncertainty: an entropy-based approach}


\author*[1]{\fnm{Danny} \sur{Wood}}\email{danny@fuzzylabs.ai}

\author[2]{\fnm{Theodore} \sur{Papamarkou}}\email{theo.papamarkou@manchester.ac.uk}

\author[1,3]{\fnm{Matt} \sur{Benatan}}\email{matt.benatan@sonos.com}

\author[1]{\fnm{Richard} \sur{Allmendinger}}\email{richard.allmendinger@manchester.ac.uk}

\affil[1]{\orgdiv{Alliance Manchester Business School}, \orgname{The University of Manchester}, \orgaddress{\city{Manchester}, \postcode{M13 9PL}, \country{UK}}}

\affil[2]{\orgdiv{Department of Mathematics}, \orgname{The University of Manchester}, \orgaddress{\city{Manchester}, \postcode{M13 9PL}, \country{UK}}}

\affil[3]{\orgname{Sonos Experience Limited}, \orgaddress{\city{London}, \postcode{SE1 3ER}, \country{UK}}}


\abstract{In order to trust the predictions of a machine learning algorithm, it is necessary to understand the factors that contribute to those predictions. In the case of probabilistic and uncertainty-aware models, it is necessary to understand not only the reasons for the predictions themselves, but also the reasons for the model's level of confidence in those predictions. 
In this paper, we show how existing methods in explainability can be extended to uncertainty-aware models and how such extensions can be used to understand the sources of uncertainty in a model's predictive distribution.
In particular, by adapting permutation feature importance, partial dependence plots, and individual conditional expectation plots, we demonstrate that novel insights into model behaviour may be obtained and that these methods can be used to measure the impact of features on both the entropy of the predictive distribution and the log-likelihood of the ground truth labels under that distribution. With experiments using both synthetic and real-world data, we demonstrate the utility of these approaches to understand both the sources of uncertainty and their impact on model performance.}




\maketitle

\section{Introduction}\label{sec:introduction}

A common criticism leveled against models in modern machine learning is that their complexity makes it difficult to understand the reasons for their decisions. This is especially problematic when applying these models in domains where decisions must be carefully justified, such as healthcare and judiciary decision making~\citep{kelly2020explainable}, making explainability an important tool when developing trustworthy AI. Another key factor in the development of trustworthy models is uncertainty quantification: the ability of models to give accurate assessments of the uncertainty inherent in their predictions. Although both explainability and uncertainty quantification are vital for the use of machine learning in high-stakes applications, remarkably little work has been done at their intersection.

Traditional explainable machine learning seeks to reveal the reasons for a model's output or the reasons for its level of accuracy. These remain important pieces of information for understanding uncertainty-aware models, though equally important is the ability to determine the sources of uncertainty: the features that cause the model to be more or less confident in its prediction. \reviewerThree{If we can understand how different features contribute to uncertainty estimates, we can engineer systems capable of better decision making. As an illustrative example, say we have a model which predicts disease activity for a patient. The model may predict a low likelihood of disease activity, but with a high associated uncertainty. Understanding the source of this uncertainty is useful in determining how to handle the model prediction; if the uncertainty is due to an unusual value of a single feature, which is known to have high sensitivity and low specificity, it may be sensible to trust the model prediction. However, if the uncertainty is due to interactions between multiple features, this may be more indicative of an out-of-distribution case in need of further exploration.} 

Although there has been some recent work exploring this area for Bayesian neural networks~\citep{depeweg2017sensitivity, Chai2018UncertaintyEI, antoran2021getting}, to the best of our knowledge, there has been only one recent attempt at a truly model-agnostic approach~\citep{watson2023explaining}.

In this paper, we demonstrate that a range of existing simple techniques for explaining model output and performance can be modified to explain both the uncertainty and likelihood of the predictive distributions of uncertainty-aware models. 
In particular, we introduce novel adaptations of permutation feature importance (PFI), partial dependence plots (PDP), and individual conditional expectation (ICE) plots to explain how each feature available to a model affects its predictive distribution. 
We explore two complementary approaches: one looking at feature importance for the negative log-likelihood (loss), the other looking at the predictive uncertainty. 

For the first approach, we introduce Likelihood-PFI, Likelihood-PDP, and Likelihood-ICE: likelihood-based variants of the methods listed above, which allow feature importance to be measured in terms of the effect a feature has on the model's negative log-likelihood. \reviewerThree{Whereas conventional PFI and PDP/ICE explain the role that given features play in determining a model's loss and output, respectively, our variants explain the role of those features in the likelihood that the model assigns to observed values of the target variable}

For the second approach, we measure the uncertainty of the model's predictive distribution through the entropy of that distribution, introducing Entropy-PFI, Entropy-PDP, and Entropy-ICE. \reviewerThree{These variants examine the role of features in determining a model's uncertainty, and can be used to reveal how features may share information and how doing so may affect a model's confidence in its predictions.}

We examine the properties of both sets of empirical measures through the use of carefully constructed synthetic datasets, and demonstrate how they can be used in a model-agnostic way to derive insights from models trained on real-world datasets in both classification and regression settings.

In recent years, methods such as those listed above have faced scrutiny due to the potential for misleading results in the presence of statistical dependencies between features~\citep{hooker2021unrestricted}. These results are due to the fact that permutation-based methods break dependencies, which can force the model to extrapolate.
While this extrapolation can be an issue for traditional permutation-based methods, we will show that this is not the case for Entropy-PFI; instead, the effect that these dependencies have on model confidence is a critical component of what is being measured.
From this interpretation, measuring Entropy-PFI alongside Likelihood-PFI can help to detect issues when dealing with feature dependencies, while not requiring training of additional models.

The remainder of the paper is structured as follows. In Section~\ref{sec:background}, we briefly review the fields of uncertainty quantification and feature importance, as well as give a more in-depth introduction to the main feature importance methods of interest in this work. In Section~\ref{sec:pfi}, we introduce our novel likelihood and entropy-based feature importance methods, as well as demonstrate their properties through experiments on synthetic datasets. In Section~\ref{sec:applied}, we demonstrate how these techniques can be used to gain new insights on real-world datasets. Finally, in Section~\ref{sec:conclusion}, we summarise our findings and give suggestions for future work.

\section{Background}\label{sec:background}

This section briefly reviews the fields of uncertainty quantification and feature importance, before introducing notation and describing some of the key methods in more detail.

\subsection{Uncertainty quantification}

Producing accurate measures of uncertainty requires that we are able to construct models that output a distribution over possible outcomes, as opposed to a single value deemed most likely by the model. To this end, several approaches have been proposed to create uncertainty-aware models, such as Gaussian processes~\citep{williams2006gaussian} and Bayesian neural networks. For Bayesian neural networks, several approaches have been developed, including fully Bayesian networks~\citep{neal2012bayesian}, approximations such as Monte Carlo dropout~\citep{gal2016dropout}), along with a range of calibration methods~\citep{guo2017calibration}. These uncertainty-aware models have their own unique strengths and weaknesses, although they all raise a set of common questions: Can we identify the sources of uncertainty for a model? And how do the features that increase model confidence differ from those that increase model performance? That is, how do we explain the uncertainty of a model?

Much previous work on explaining uncertainty revolves around decomposing the uncertainty into two types: \emph{epistemic} uncertainty, i.e., uncertainty due to the finite amount of data available and limitations of the model; and \emph{aleatoric} uncertainty, i.e., uncertainty that is inherent in the system that we are observing, which cannot be reduced by collecting more data~\citep{depeweg2018decomposition}. Epistemic uncertainty can be further decomposed into uncertainty about the suitability of the chosen model (structural uncertainty) and uncertainty in the choice of parameters given the specification of the chosen model~\citep{liu2019accurate}. Various efforts have been made to identify how much uncertainty comes from each source: explicitly modelling aleatoric uncertainty via a noise parameter, estimating the role of epistemic uncertainty by modelling the density of the data in latent spaces~\citep{mukhoti2023deep} and estimating aleatoric uncertainty by reducing epistemic uncertainty via ensembling~\citep{shaker2020aleatoric}, while other work critically examines the validity of this approach of decomposing uncertainty~\citep{wimmer2023quantifying}.

However, we are interested in more fine-grained explanations for the sources of uncertainty: explaining how the particular values of given features in an example affect the uncertainty of the model output.
Little work has been done in this area, although there are notable exceptions.
\cite{depeweg2017sensitivity} examined the role of features in the uncertainty of Bayesian neural networks through sensitivity analysis: examining the gradient of the uncertainty with respect to each feature in turn.
\cite{antoran2021getting} look at how uncertainty estimates can be explained in Bayesian neural networks via counterfactuals: identifying which constellations of features are responsible for uncertainty by finding sets of minimal changes required to increase a model's confidence in its prediction. 
Similarly,~\cite{Chai2018UncertaintyEI} and~\cite{zhang2022explainable} look at the importance of features through predictive difference, showing changes in the predictive distribution of Bayesian neural networks when features are replaced with non-informative features modelled from a conditional distribution. 
More recently,~\cite{watson2023explaining} have considered how Shapley values could be used to explain both aleatoric and epistemic uncertainty in a completely model-agnostic manner. This allowed for local explanations of the behaviour of model uncertainty.

Our work shares the same motivation as these works: to explain not just the predictions of these models, but also the uncertainty associated with those predictions.
Unlike~\cite{depeweg2017sensitivity, Chai2018UncertaintyEI, zhang2022explainable} and~\cite{antoran2021getting}, we do not assume a particular structure for our model and data. Although similar in spirit to the work of~\cite{watson2023explaining}, our work uses techniques that are conceptually and computationally simpler, while still able to offer insight into model behaviour.

\subsection{Feature importance}

Early feature importance methods were developed in an ad hoc way to describe models of interest, rather than being the subjects of research in their own right. Permutation feature importance (PFI) and partial dependence plots (PDPs) were first introduced in articles on random forests~\citep{breiman2001random} and gradient boosting~\citep{friedman2001greedy}, respectively. Although in both cases, the advantages and disadvantages of the feature importance measures were discussed, such a discussion was secondary to the main objectives of the article. Despite this, both methods have been widely adopted, adapted, and extended in the literature.
Most notably for our purposes,~\cite{moosbauer2021explaining} introduced the idea of including confidence bands on PDPs to measure the uncertainty associated with a cost function when applying PDPs to a cost function in Bayesian optimisation.


Although PFI and PDPs were introduced independently, they share many common features. Implicit in both is the assumption that we can break the dependence of the target variable from the feature of interest by sampling from that feature's marginal distribution without generating an out-of-distribution sample. However, this is not a reasonable assumption---as was indeed observed in different respects in the original papers introducing the two measures~\citep{breiman2001random, friedman2001greedy}.

These shortcomings have been addressed for PFI using conditional methods~\citep{strobl2008conditional, molnar2023model}. In~\cite{strobl2008conditional}, the issue is resolved for random forests by taking advantage of the tree structure to break the feature space into a grid, with permutations carried out only between features which share a common region, allowing for an approximation of the conditional distribution. This issue is addressed in a model-agnostic way in~\citep{molnar2023model}: by splitting the test set into sub-groups in which the features of interest are conditionally independent of the other features using an additional model, then performing PFI with permutations restricted to only swap values for examples within the same sub-group. An alternative approach is to compare the model of interest with a second model in which the feature of interest has been completely removed, or the information in that feature has been destroyed by permutation in the training set~\citep{hooker2021unrestricted}. These approaches are referred to as remove-and-relearn and permute-and-relearn, respectively. 
These approaches resolve many of the issues that permutation-based methods face but do so at greater computational cost, requiring at least one more model to be trained for each feature of interest.

For PDP, the issue of extrapolation can be partially resolved through the use of individual conditional expectation (ICE) plots~\citep{goldstein2015peeking}. These de-aggregate the effects of individual test examples on PDP, allowing users to see how examples differ in their sensitivity to a particular feature. In this way, ICE plots can be used to reveal heterogeneity in the model behaviour for a given feature.

Another popular approach for local explanations is local interpretable model-agnostic explanations (LIME)~\citep{ribeiro2016should}, in which a simple, easy-to-interpret model is used to approximate the more complex model of interest within a small region of interest. This local model is trained explicitly so that it is locally a good approximation of the target model, and therefore can be analysed to determine what factors affected the model's decision at that point.

The SHAP algorithm is a method to derive point-wise explanations for a model's output in a principled manner~\citep{lundberg2017shap}. SHAP is derived from the game-theoretic notion of Shapley values, and is provably unique in satisfying a specific set of desirable properties~\citep{lundberg2017shap}. Although SHAP values are theoretically sound, they suffer from problems of tractability. Several approximations are suggested by~\cite{lundberg2017shap}, and there are model-specific variants that allow for more accurate calculation~\citep{lundberg2020treeshort}, as well as model specific variants incorporating uncertainty~\cite{chau2024explaining}. 
A detailed discussion on the ways in which Shapley values can be computed and estimated can be found in~\citep{chen2023algorithms}.
Like permutation-based methods, both LIME and SHAP suffer when forced to extrapolate. For these methods, this difficulty comes in the form of adversarial attacks, which have been shown to allow a malicious actor to create false/misleading explanations by creating perturbed examples that are separate from the true data distribution~\citep{slack2020fooling}.

Recently, a common framework unifying many of the methods described above, amongst others, was developed under the paradigm of ``explaining by removing''~\citep{covert2021explaining}. Using this framework,~\cite{covert2021explaining} conducted a systematic exploration of the connections and differences between existing model explanation methods.

\subsection{Permutation-based feature importance methods}

In this section, we describe the key feature importance methods from the literature in more detail. 
These methods provide a tool for explaining how much a particular feature's value is responsible for determining the performance of a model.

\subsubsection{Permutation feature importance (PFI)}\label{subsec:pfi}

Define the random variables for the feature vector and the target as $X$ and $Y$, respectively, and let $P_{XY}$ denote the true joint distribution of the data, with $(X, Y) \sim P_{XY} $. 
Furthermore, let $P_X$ denote the marginal distribution of $X$. We write $X = (X_1, \ldots, X_d)$, with $X_j$ being the random variable corresponding to the $j$-th feature, and denote its marginal distribution $P_{X_j}$. 
Furthermore, we define $X_{-j}$ as the vector of features with the $j$-th element omitted, that is, $X_{-j}=(X_1, \ldots, X_{j-1}, X_{j+1},\ldots,X_d)$, similarly denoting its joint distribution as $P_{X_{-j}}$.
Throughout, we use the convention that $i$ indexes examples in a test set (with $1 \leq i \leq n $), $j$ indexes features in a feature vector ($1 \leq j \leq d$) and $c$ indexes class labels ($1 \leq c \leq k$).

PFI works as follows: given a trained model and a set of test examples $\{(\vectorx^{(1)}, y^{(1)}),\ldots, (\vectorx^{(n)}, y^{(n)})\}$, with each $\vectorx^{(i)} \in \mathbb{R}^d$, we construct a design matrix in  $\mathbb{R}^{n \times d}$, where the $i$-th row is the transpose of $\vectorx^{(i)}$. To obtain the PFI measurement for the $j$-th feature, we randomly permute the $j$-th column and use the rows with that feature permuted as the feature vectors for our new test set. We refer to this new test set with the notation $\{(\widetilde{\vectorx}^{(1)}, y^{(1)}),\ldots,(\widetilde{\vectorx}^{(n)}, y^{(n)})\}$. With a cost function $\loss$, the empirical PFI for this test set is given as
\begin{align}
    \empiricalPFI(j) = \frac{1}{n}\sum_{i=1}^n 
    \loss(y^{(i)}, f(\vectorx_{-j}^{(i)}, \widetilde{x}_j^{(i)}))  - \loss(y^{(i)}, f(\vectorx_j^{(i)})).\label{eq:emp_pfi}
\end{align}
Here, we allow ourselves to use the convention adapted from~\cite{casalicchio2019visualizing} of writing $f(\xnotj, \xcopy_j)$ to mean the function $f$ with an input vector where the $j$-th entry is $\xcopy_j$ (i.e., the $j$-th entry of $\widetilde{\vectorx}$) and the other entries are populated using the entries of $\xnotj$.
We may think of the PFI for the $j$-th feature as the difference in performance when the dependence of the model on the $j$-th feature is broken by permuting the $j$-th feature in the test examples.

The values obtained by PFI are specific to the test set considered. However, they are Monte Carlo approximations of a quantity defined in terms of the data distribution~\citep{casalicchio2019visualizing}. 
Consider $X, Y \sim P_{XY}$, with $\Xnotj$ being a subset of the features of $X$. Let $\Xcopy_j$ have the same marginal distribution as $X_j$ but independent of $X$ and $Y$. We have that
\begin{align}
   \expectedPFI(j) &= \mathbb{E}_{\Xnotj, \Xcopy_j}[\mathcal{C}(Y, f(\Xnotj, \Xcopy_j))] - \mathbb{E}_{X}[\mathcal{C}(y, f(X))]\label{eq:exp_pfi}\\
   &\approx \frac{1}{n} \sum_{i=1}^n \mathcal{C}(y^{(i)}, f(\xnotj^{(i)}, \xcopy_j^{(i)} )) - \mathcal{C}(y^{(i)}, f(\vectorx^{(i)}))\nonumber\\
   &= \empiricalPFI(j)\nonumber.
\end{align}


A common modification of PFI is to condition the distribution of the feature of interest on the observed values of the other features. This way, conditional PFI (CPFI) is defined as 
\begin{align*}
   \expectedcPFI(j) = \mathbb{E}_{\Xnotj, (\widetilde{X}^c_j|\Xnotj)}[\loss(Y, f(\Xnotj, \Xcopy^c_j))] - \mathbb{E}_{X}[\loss(y, f(X))],
\end{align*}
where $\widetilde{X}^c_j$ is constructed such that it follows the conditional distribution of $X_j|\Xnotj$ but $\widetilde{X}^c_j|\Xnotj$ is independent of $Y$~\citep{strobl2008conditional}. 

\subsubsection{Partial dependence plots (PDPs)}\label{subsec:pdp}

While PFI is a measure of the effect of a given feature on the model's performance, given a ground-truth label, PDP gives a method of visualising a feature's effect on the model output itself.
In PDP, a single feature is kept constant, while all other features jointly assume values from an example in the test set; the average model output is then captured across the test set. For feature $j$, the PDP is found by plotting $\empiricalPDP(x_j; j)$ for all values $x_j$, where $\empiricalPDP$ is defined as
\begin{align}
   \empiricalPDP(x_j; j) = \frac{1}{n} \sum_{i=1}^n f(\xnotj^{(i)}, x_j),\label{eq:pdp_def}
\end{align}
This is a Monte-Carlo approximation of the true value of interest, that being
\begin{align*}
    \expectedPDP(x_j; j) = \mathbb{E}_\Xnotj [f(\Xnotj, x_j)].
\end{align*}
\reviewerOne{The value of a PDP at point $x_j$ can be thought of as the answer to the following question: given that all features except the $j$-th follow the joint data distribution, what is the expected value of the model output when fixing the $j$-th feature to the value $x_j$. Note that like PFI, in computing this value, the model can be forced to extrapolate to sets of features that do not occur in the true data distribution.}

An example of the kinds of curve produced by PDP is shown in Figure~\ref{fig:pdp_ice_demo}. In this demonstration, we show the PDP curve for a single hidden layer multi-layer perceptron (MLP) on the diabetes dataset~\citep{smith1988using}.

\subsubsection{Individual conditional expectations (ICEs)}\label{subsec:ice}

A PDP is most effective when the distributions of features are independent; as shown by~\cite{goldstein2015peeking}, a PDP can hide the real effects of varying a feature by considering only the average over the training distribution, rather than examining the effects on individual examples. To this end,~\cite{goldstein2015peeking} introduced individual conditional expectation (ICE) plots, which show how the model outputs for individual examples change as the feature of interest is varied. For feature of interest $j$, the ICE plot for the test example $i$ is given by plotting the function
\begin{align}
   \ICE^{(i)} (x_j; j) = f(\xnotj^{(i)}, x_j).\label{eq:ice_def}
\end{align}
The PDP curve is simply the average of ICE curves over the test set, as can be trivially observed from Equations~\ref{eq:pdp_def} and~\ref{eq:ice_def}.

\begin{figure}[ht]
    \centering
    \includegraphics[width=.7\textwidth]{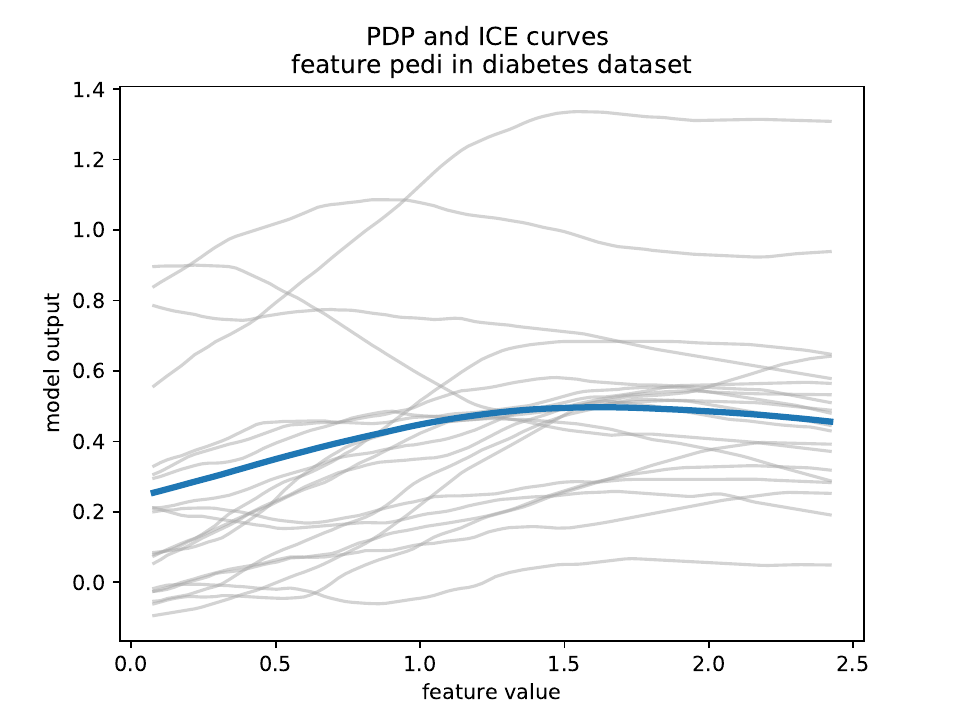}
    \caption{Example of PDP and ICE plots. The blue line shows the PDP. The ICE curves (gray) show how the output of the model changes for individual examples as the feature of interest changes.}
    \label{fig:pdp_ice_demo}
\end{figure}

\reviewerOne{An individual ICE curve shows how the model output for a training example would vary if we were to alter the value of the  $j$-th feature. By plotting the ICE curve over the range of values the $j$-th feature takes in the data, we observe how the value of the $j$-th feature affects the model's prediction for the given sample. In Figure~\ref{fig:pdp_ice_demo}, we see that while the PDP curve shows the average behaviour, individual ICE curves show that the effects on model output of change a single feature can vary significantly depending upon the other feature's values.}

\section{Explaining likelihood and uncertainty}\label{sec:pfi}

In this section, we propose modifications to the feature importance methods described in Sections~\ref{subsec:pfi} and \ref{subsec:pdp} to capture explanations for the likelihood and uncertainty of the predictive distribution of uncertainty-aware models.

\subsection{Likelihood-PFI}

We begin with PFI for negative log-likelihood, which we refer to as Likelihood-PFI: this is a natural extension of the original PFI measure, with the only difference being that we exchange the loss function used in the traditional PFI setting with the negative log-likelihood of the target given the feature variables. Rather than defining a model output $f$, we now think of the model as giving a predictive \emph{distribution} $q$ and use the negative log-likelihood of the target given this distribution as our loss function. 

Given a test set $\{(\vectorx^{(1)}, y^{(1)}), \ldots, (\vectorx^{(n)}, y^{(n)})\}$ of size $n$, the empirical Likelihood-PFI is given by 
\begin{align}
    \empiricalLPFI(j) =  \frac{1}{n} \sum_{i=1}^n \log q(y^{(i)} | X=\vectorx^{(i)}) - \log q(y^{(i)} | \Xnotj=\xnotj^{(i)}, \Xcopy_j=\widetilde{x}^{(i)}_j  ).\label{eq:emp_pfi_like_def}
\end{align}
In order to obtain the exact value of which this is an approximation, we write
\begin{align}
    \expectedLPFI(j) =  \mathbb{E}_{X, Y} [ \log q(Y | X)] - \mathbb{E}_{\Xcopy_j, \Xnotj, Y}[\log q(Y|\Xnotj, X_j=\Xcopy_j)], \label{eq:exp_pfi_like_def}
\end{align}
where the $X_j=\Xcopy_j$ in $q(Y|\Xnotj, X_j=\Xcopy_j)$ denotes the fact that while under the model distribution, Y is conditional on $X_j$, not $\Xcopy_j$, when we take the expectation, we treat $X_j$ as being distributed according to the marginal, rather than the joint distribution. 
This is clarified further in Appendix~\ref{app:regarding_notation}.

Note that the order of terms in Equations~\ref{eq:emp_pfi_like_def} and \ref{eq:exp_pfi_like_def} is reversed in comparison to the terms in the original PFI definitions (Equations~\ref{eq:emp_pfi} and \ref{eq:exp_pfi}). This is because the function  under consideration here is the \emph{negative} log-likelihood, and reversing the order of terms allows us to avoid a double negative. However, the quantity of interest is still the loss given the permuted feature minus the loss given the original feature value. 

For the regression case, the Likelihood-PFI is related to the PFI for deterministic models; for instance, given a fixed-width Gaussian predictive distribution, the Likelihood-PFI reduces to the PFI of the squared loss between the predictive mean and the observed value of the target variable (up to a constant scaling factor). For the classification case, PFI is typically performed on either the misclassification rate~\cite{breiman2001random} or the AUC \citep{molnar2022interpretable}, whereas our approach would reduce to performing PFI on the log-loss. Unlike using the misclassification rate, this approach allows for detection of small changes in the model's confidence that are not significant enough to change the class prediction, and unlike AUC it generalises naturally from binary classification to a multi-class setting.
\reviewerThree{Like traditional PFI, we expect Likelihood-PFI to be non-negative: removing information from an informative feature will increase the loss, whilst permuting uninformative features will have no effect.}

\subsection{Entropy-PFI}

Likelihood-PFI gives a measure of how the model \emph{performance} is affected by re-sampling a feature from its marginal distribution. However, it is also useful to know how the \emph{uncertainty} of a model is affected by the value of each feature.
To this end, we next look at how PFI can be performed for the entropy of its prediction rather than for its accuracy.

 The Shannon entropy for a given categorical random variable $Y$ taking values in $\mathcal{Y}=\{1, \ldots, k\}$, given a model distribution $q$, is given by
\begin{align*}
   \entropy(Y) = - \sum_{y \in \mathcal{Y}} q(y) \log q(y),
\end{align*}
where the $q$ subscript in $\entropy$ emphasises that we consider the entropy under the model's predictive distribution (in contrast with the entropy of the true predictive posterior). Similarly, for a continuous random variable, the entropy is written as
\begin{align*}
    \entropy(y) = - \int q(y) \log q(y) \,  \,dy.
\end{align*}
With this, we define the entropy permutation feature importance (Entropy-PFI) as
\begin{align*}
    \expectedEPFI(j) &= \mathbb{E}_{X, \widetilde{X}_j}\left[ \entropy(Y|\Xnotj, \Xj=\Xcopy_j) - \entropy(Y| X)\right],
\end{align*}
and we get the Monte-Carlo approximation using a test set as 
\begin{align*}
    \empiricalEPFI(j) &= \frac{1}{n} \sum_{i=1}^n \entropy(Y|\Xnotj=\xnotj^{(i)}, \Xj=\widetilde{x}_j) - \entropy(Y| X=\vectorx^{(i)}).
\end{align*}
\reviewerOne{We note that in the continuous case, the entropy may be negative as well as positive. For the regression models we consider, the predictive distribution will be Gaussian, and the entropy of that distribution may be written as }
\begin{align*}
    \mathcal{H}_q(y) = \frac{1}{2} + \frac{1}{2} \log (2 \pi \sigma^2),
\end{align*}
\reviewerOne{where $\sigma^2$ is the variance of the distribution. This value is negative when $\sigma^2$ is sufficiently small.}

We can think of Entropy-PFI as measuring how much uncertainty increases on average when we replace the $j$-th feature of an example with a random sample from its marginal distribution. Intuitively, we would expect for this value to be non-negative: by replacing this feature, we will often be moving away from dense regions in the sample space, where the model will have low epistemic uncertainty, to sparser ones, where the model will have seen fewer examples and therefore exhibit lower confidence in its predictions. So, \emph{a high PFI score means that the value of a feature helps to increase the confidence of the model in its prediction.}
Note that, unlike the Likelihood-PFI, the Entropy-PFI is independent of the ground-truth label.

\subsection{Properties of Entropy-PFI}

In interpreting Entropy-PFI, it is important to understand exactly what the quantity measures. For a given feature, it gives a measure of how much the value of that feature supports the model's conclusion derived from the other features. If the feature shares task-relevant information with other features, the model will be more confident when the feature under consideration agrees with those features and less confident when the relationship between the feature values is destroyed. Entropy-PFI establishes the difference between these two levels of confidence by comparing the level of confidence under the true distribution against the level of confidence when the feature of interest follows the same marginal distribution, but is independent of the other features.

\begin{figure}[ht]
    \centering
    \includegraphics[width=\textwidth]{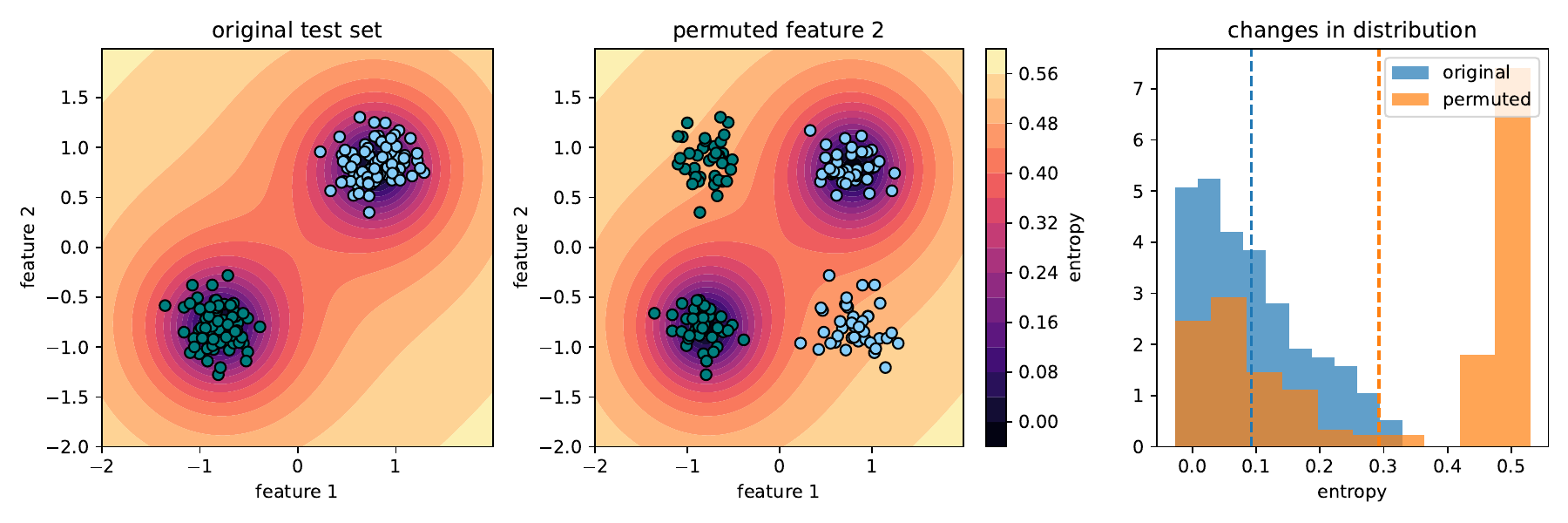}
    \caption{Visualisation of effects of PFI. The colour of each point shows the cluster to which the original (unpermuted) test example belonged. In the left panel, the original test set is shown, along with the contour lines for the entropy of a (hypothetical) model's predictive distribution. In the centre panel, the test set is shown after permuting feature 2. In the right panel, histograms of the entropy before and after permuting the second feature are shown.}
    \label{fig:property_1}
\end{figure}

In Figure~\ref{fig:property_1}, we see a mock-up of how permuting features affects the entropy of a test set by moving examples from low-entropy regions to high-entropy ones. In the left panel, the contour plot shows the values of the entropy over the feature space, and each dot represents a member of the test set. We see that the test examples all occur in low-entropy areas: the combined information of the two features allows the model to be certain in its prediction. In the centre panel, we perform PFI on the second feature. Here, we see that there are test points which are in high-entropy areas: we can think of this as the model being surprised by the combination of features, and increasing its level of uncertainty as a result. On the right, we see the histogram of the entropies. We see that when permuted, there are now more high-entropy points, and the average has increased. 

Intuitively, we can think of Entropy-PFI as answering the question ``How much does the true value of this feature support the given prediction on the basis of the evidence provided by the other features?''. A consequence of this interpretation is that we would expect that if a feature did not share any information with other features, the Entropy-PFI would be zero. In the following proposition, we verify that this is the case.

\begin{proposition}\label{prop:independent_pfi}
If $\Xnotj$ is independent of $\Xj$, then the Entropy-PFI is zero.
\end{proposition}
\begin{proof}
Starting from the definition of $\expectedEPFI$, we use the fact that $X_j$ is independent of $\Xnotj$ to swap it with $\Xcopy_j$:
\begin{align*}
    \expectedEPFI(j) &= \mathbb{E}_X \left[ \mathbb{E}_{\Xcopy_j} [\entropy(Y|\Xnotj, X_j=\Xcopy_j)] - \entropy(Y|X)  \right] \\
     &= \mathbb{E}_X \left[ \mathbb{E}_{X_j} [\entropy(Y|\Xnotj, X_j)] - \entropy(Y|X)  \right] \\
     &=\mathbb{E}_X \left[  \entropy(Y|X) - \entropy(Y|X)  \right]=0 .
\end{align*}
\end{proof}

We note that this is not the case for the Likelihood-PFI, or the PFI of any target-dependent measure in general, where dependencies between $\Xj$ and $Y$ prevent the substitution used in the proof above. \reviewerOne{Intuitively, the result is caused by the fact that if a feature is independent, resampling that feature (e.g., by swapping its value with another from the same test set) will generate another sample from the same underlying data distribution.}
The above proposition means that when a feature is independent of the others, it does not globally affect the confidence of the model. However, this does not mean that the feature does not affect the model's confidence locally, only that local effects cancel out in aggregate.

A feature being independent of the complementary set is not the only way that Entropy-PFI can be zero. It can also be zero if the predictive distribution does not depend on the feature of interest. This is analogous to how traditional PFI will be zero if a feature is not used in determining the model output.

\begin{proposition}\label{prop:not_relevant}
    If the predictive distribution is not dependent on feature $j$, i.e., $q(Y|X) = q(Y|\Xnotj)$, then $\expectedEPFI(j)=0$ and $\expectedLPFI(j)=0$ .
\end{proposition}
\begin{proof}
   By assumption, and by definition of $\entropy$, we have that $\entropy(Y|\Xnotj, X_j) = \entropy(Y|\Xnotj)$. By definition of $\expectedEPFI$, we therefore have
   \begin{align*}
    \expectedEPFI(j) &= \mathbb{E}_X \left[ \mathbb{E}_{\Xcopy_j} [\entropy(Y|\Xnotj, X_j=\Xcopy_j)] - \entropy(Y|X)  \right] \\
    &= \mathbb{E}_X \left[\entropy(Y|\Xnotj) - \entropy(Y|\Xnotj)  \right] =0 .
   \end{align*}
   Similar reasoning gives the result for $\expectedLPFI(j)$:
      \begin{align*}
    \expectedLPFI(j) &= \mathbb{E}_{X,Y} \left[ q(Y|X)\log q(Y|X) - \mathbb{E}_{\Xcopy_j} [q(Y|\Xnotj, X_j=\Xcopy_j) \log q(Y|\Xnotj, X_j=\Xcopy_j)]   \right] \\
    &= \mathbb{E}_{X, Y} \left[[q(Y|\Xnotj)\log q(Y|\Xnotj) - q(Y|\Xnotj)\log q(Y|\Xnotj)  \right] =0 .
    \end{align*}
\end{proof}

\reviewerOne{It is worth noting that both of the above propositions apply to computing the over the data distribution. In practice, the empirical Entropy-PFI will be non-zero, but its typical deviation from zero will decrease with test set size.}

\subsection{Why conditional PFI is not useful in the context of entropy}

At their core, permutation-based feature importance methods rely on re-sampling features based on their marginal distribution, breaking the relationship between the feature and the set of all other variables. However, doing this leads to undesirable outcomes: by ignoring correlations and other relationships between feature variables, we can find ourselves evaluating the model on points outside the true data distribution.  
As discussed in depth by~\cite{hooker2021unrestricted}, this can be problematic in that the resulting feature importances rely on the extrapolating behaviour of the model, giving importance measures that are dependent on the model's behaviour in regions far from the training data, where the model's behaviour is unlikely to reflect the true distribution of the data.

One of the proposed methods for dealing with this is through conditional approaches, where a variable $\widetilde{X}^c_j$ is constructed so that it retains the same conditional relationship with $\Xnotj$ as $X_j$, but is independent of $Y$ given the information contained in $X_j$.  However, such approaches are at odds with what Entropy-PFI measures: Entropy-PFI is non-zero only when features share task-dependent information and permuting one of the features breaks the dependency between one feature and the set of other features.

Say that we have access to a random variable $\Xjcond$ such that $(\Xjcond, \Xnotj)$ has the same joint distribution as $(\Xj, \Xnotj)$ but $\Xjcond$ is independent of $Y$ (either completely or when conditioned on $\Xnotj$). We can define Likelihood-PFI in the same way that conventional PFI is defined for this approach. However, if we attempt to define conditional Entropy-PFI in the same way, e.g., 
\begin{align}
    \expectedcEPFI(j)
     =  \mathbb{E}_{X, \Xjcond}[\entropy(Y|\Xnotj, X_j=\Xjcond) - \entropy(Y|X)]. \label{eq:cond_pfi_def}
\end{align}
we cannot use it to get estimates of the uncertainty caused by feature $X_j$, since this will always be zero, as shown in the following proposition.

\begin{proposition}
The entropy version of conditional PFI, as defined in Equation~\eqref{eq:cond_pfi_def}, is zero for all features.
\end{proposition}
\begin{proof}
Considering an arbitrary feature indexed by $j$, we look at the first term in more detail, finding
\begin{align*}
    \mathbb{E}_{X, \Xjcond}[\entropy(Y|\Xnotj, X_j=\Xjcond)] = \mathbb{E}_{X, \Xjcond}\left[\int u(y|\Xnotj, X_j=\Xjcond) dy\right],
\end{align*}
where $u(\cdot) = q(\cdot)\log q(\cdot)$ for the sake of brevity.
Using the fact that $(\Xjcond, \Xnotj)$ has the same joint distribution as $(\Xj, \Xnotj)$ by definition, we can expand the expectations to give
\begin{align*}
    \mathbb{E}_{X, \Xjcond}&[\entropy(Y|\Xnotj, X_j=\Xjcond)]\\
    =&-\int \int \int u(y| \Xnotj=\xnotj, X_j=\widetilde{x}_j) P_{\Xnotj, \Xjcond}(\xnotj, \widetilde {x}_j) \,dy d\vectorx_{-j} dx_j \\
    =&-\int \int \int u(y| \Xnotj=\xnotj, X_j=\widetilde{x}_j) P_{\Xnotj, \Xj}(\xnotj, \widetilde {x}_j) \,dy d\vectorx_{-j} dx_j\\
    =&-\int u(y| X=\vectorx) \,dy\, P_{X}(\vectorx) d\vectorx\\
    =& \mathbb{E}_{X} \left[\entropy(Y|X) \right] .
\end{align*}
Plugging this into Equation~\eqref{eq:cond_pfi_def} gives that the conditional Entropy-PFI is zero.
\end{proof}
Therefore, we see that, by definition, conditional approaches for PFI are ineffective in measuring the importance of features in determining entropy.
However, we make two arguments as to why this is not problematic. Firstly, one of the ways in which Entropy-PFI has utility is in identifying when shared information between features has the effect of boosting model confidence; in attempting to preserve shared information between the feature of interest and the other features via conditioning during permutation, we eliminate the very discrepancy that we aim to measure.
Secondly, while relying on the extrapolation behaviour of a model is in general undesirable, especially when the underlying data generating process is of interest, for an uncertainty-aware model, its ability to display increased (epistemic) uncertainty when extrapolating is one of the key desirable characteristics of the model. 

\subsection{How to interpret Entropy-PFI}

Given that Entropy-PFI becomes zero when a feature is independent of the others or when we construct conditional variants of it, it is clear that its behaviour is not exactly analogous to that of traditional PFI. In contrast, Likelihood-PFI can be thought of as a direct port of the traditional loss-based PFI to an uncertainty-aware setting, inheriting the known properties and issues from PFI itself.

\reviewerThree{As such, it is worth spending some time discussing exactly how Entropy-PFI should be interpreted and how it can be used to inform us about our model before we attempt to use the measure in practice. In this section, we discuss how Entropy-PFI can be used in conjunction with Likelihood-PFI to derive insights from the data, as well as overcome some of Likelihood-PFI's shortcomings.}

\paragraph{When Entropy-PFI is zero} \reviewerFour{In Proposition~\ref{prop:independent_pfi} and Proposition~\ref{prop:not_relevant}, we gave two settings in which the Entropy-PFI is zero: one in which the feature is informative to the model but independent of other features, and the other in which the feature does not contain any information used in determining the predictive distribution. These are two cases that it is important to differentiate between, so it is reasonable to ask whether this means Entropy-PFI is too weak a tool to truly explain model behaviour. }

\reviewerFour{Though Entropy-PFI in isolation does not allow us to discern the difference between the two scenarios, in both these cases, Entropy-PFI still serves as a useful complement to Likelihood-PFI to explain the role of the given feature in the model's prediction. If the Likelihood-PFI is also zero, this suggests that the feature is uninformative, as changing its value does not have an impact on model performance. On the other hand, if Likelihood-PFI is positive, it means that the feature is informative, but the information it contains about the target variable is not shared with other features.}

\paragraph{When Entropy-PFI is non-zero} 
\reviewerFour{In the case where the Entropy-PFI is non-zero, this means that a) the feature contains information about the target variables and b) some (or all) of that information is also contained jointly in the other features. From~\citet{molnar2022interpretable}, we know that if information from a feature is available from other sources, the importance of that feature as measured by PFI will be diminished. As such, when we see a high Entropy-PFI, we should expect that the Likelihood-PFI under-reports the utility of the information contained in that variable, since the model will sometimes be extracting that information from a different source.}

\paragraph{The role of extrapolation}
\reviewerThree{Entropy-PFI measures the difference of a quantity in expectation over the marginal and the joint distributions of the feature of interest (and in expectation over the joint distribution of all the other features). For this to be non-zero, these distributions need to be different, and the marginal distribution will necessarily put more weight on some areas which are low/zero density in the joint distribution. For this reason, the measure can be thought of as inherently dependent on how a model determines its confidence levels in these regions. In an extreme case where the model gives the same uncertainty estimate everywhere, the Entropy-PFI will be zero for all features, and Likelihood-PFI reduces to PFI. }

\reviewerThree{However, uncertainty-aware models are, in general, designed to give higher uncertainty estimates in regions where they have been presented with little data. This scenario is modeled in Figure~\ref{fig:property_1}, with the increase in uncertainty being caused by the model being less confident in out-of-distribution regions. Under this assumption, we can think of the uncertainty as being largely epistemic in nature: where training data is sparser, uncertainty-aware models will exhibit higher uncertainty. Additionally, it is under this assumption that we expect Entropy-PFI to be non-negative: a model will likely be more confident interpolating than extrapolating. However, we also note that model behaviour will vary even between uncertainty-aware models, depending on their extrapolation behaviours and this variation is an additional factor to consider in interpreting model behaviours with regards to uncertainty. While these tools are model-agnostic, they require practitioners to be aware of a model's behaviour in order to correctly interpret.}

\paragraph{What kind of uncertainty do we measure?}
The mechanism by which Entropy-PFI works can be most easily understood in terms of epistemic uncertainty, but it is worth noting that it does not mean that Entropy-PFI is \emph{just} measuring epistemic uncertainty. \reviewerOne{Entropy-PFI explicitly measures the uncertainty of the predictive distribution, regardless of whether that role is epistemic or aleatoric in nature.} Given the role of extrapolation in calculation of Entropy-PFI, epistemic uncertainty will likely be significant component, but the model's understanding of aleatoric uncertainty in different regions of the feature space will also affect the importance of features.

\reviewerFour{Moreover, Entropy-PFI is dealing with the predictive uncertainty of the model over the whole feature space. While thinking in terms of epistemic and aleatoric uncertainty is useful in reasoning about a model's behaviour and interpreting the behaviour of our measures, it is important to consider that a model's understanding of these sources of uncertainty will likely be flawed/incomplete. 
In the notable case where the model has a built-in assumption of homoscedastic noise, then Entropy-PFI will indeed be measuring the model's understanding of its uncertainty due to lack of data. However, in general, the model's understanding of how aleatoric uncertainty changes over the joint feature distribution will play a role in measuring the uncertainty.
As such, while we can use our understanding of how models deal with epistemic and aleatoric uncertainty to interpret the results, we should not rely on our measures as a tool to distinguish between these two sources of uncertainty. Instead, we may use our measures as a tool operating on the uncertainty of the overall predictive distribution.}

\subsection{Examples of joint usage of Entropy-PFI and Likelihood-PFI}

We now present two examples of synthetic datasets in which Entropy-PFI and Likelihood-PFI together can enhance interpretability in terms of feature importance and predictive uncertainty. In particular, one classification and one regression example are presented.

\paragraph{Classification experiment on synthetic data} In order to examine the interpretation of Entropy-PFI and Likelihood-PFI in a classification setting, we consider a toy binary classification dataset simulated according to the model
\begin{align*}
    P(Y=1 | \vectorx) = \epsilon + (1-2\epsilon)\,\mathds{1}\left(\sum_{j=1}^J \vectorx_{j} > \frac{J}{2}\right),
\end{align*}
where $\vectorx$ is sampled uniformly from the unit hypercube $[0,1]^d$~\citep{mease2008evidence}. Here, $\epsilon$ is the amount of label noise, $d$ is the total number of features, and $J$ is the number of relevant features. For our experiments, we set $d=10$, $J=4$ and $\epsilon=0.1$. This means that there are four features all with equal importance (features 1, 2, 3 and 4) and six features which are irrelevant in determining the target class. 
We consider three versions of this data: one that is exactly as described above, a second in which feature 10 is replaced with a copy of feature 1, and a third in which feature 10 is replaced with a copy of feature 5.

\begin{figure}[ht]
    \centering
    \includegraphics[width=\textwidth]{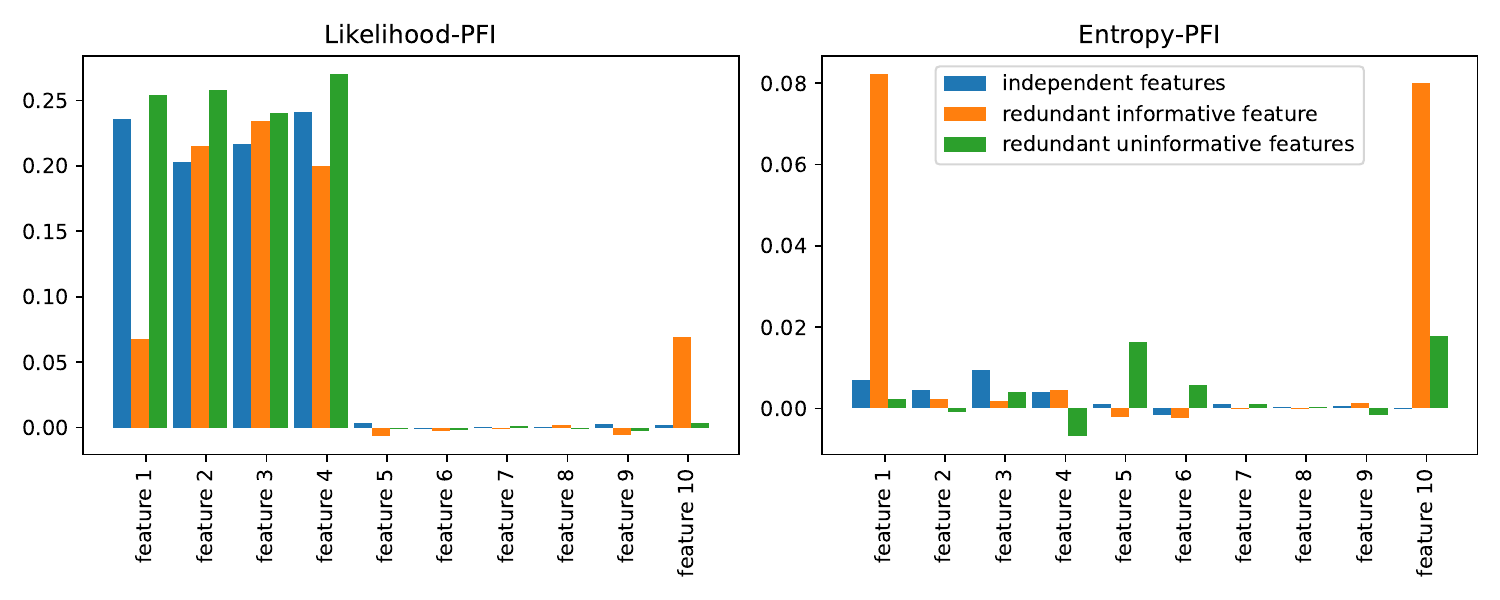}
    \caption{Comparison of Likelihood-PFI and Entropy-PFI  for three datasets, the second and third of which contain redundant features. When feature 10 is a copy of feature 1 (an informative feature), we see PFI-likelihood of feature 1 drop and PFI-entropy increase, and both PFI-likelihood and PFI-entropy increase for feature 10. When feature 10 is a copy of feature 5 (an uninformative feature), there is no effect on PFI-likelihood for either feature, and a small increase in PFI-entropy for both.   }
    \label{fig:redundant_error}
\end{figure}

For each of the three generation methods discussed above, we generate a dataset of 5000 examples with a train/test split of 3750/1250. \reviewerOne{A separate calibrated random forest is trained on the training set for each of the three datasets, then Likelihood-PFI and Entropy-PFI is computed for each on their respective test sets.}
In Figure~\ref{fig:redundant_error}, we see how the Likelihood-PFI and Entropy-PFI are affected by adding redundancy to the dataset for a random forest with calibration (see Appendix~\ref{sec:hyperparams} for details). 

In the original dataset, all four task-relevant features have similarly high Likelihood-PFI values. 
However, when feature 10 is replaced with a copy of feature 1, the first feature is now ranked as less important; this is because there is now an alternate source from which the model can get the same information.

In contrast, features 1 and 10 are the features that Entropy-PFI identifies as most important in making the model confident in its output, as shown in the right panel of the figure. This makes sense: in all the training data, these features have been strongly correlated (identical, in fact), and therefore examples where this relationship is broken should be treated as out-of-distribution, which should be reflected in greater predictive uncertainty.

We also see a small increase in Entropy-PFI for the redundant features in the third dataset. This is likely due to the fact that, despite features 5 and 10 not containing any information about the target, spurious correlations in the training set may cause the model to use these features, and therefore the model is able to identify when it goes out-of-distribution due to disagreement between the two values, resulting in changes in entropy. However, we note that this effect is small in comparison to the effect in features that are informative and therefore are actually useful  to the model. Indeed, as shown in Proposition~\ref{prop:not_relevant}, if the model (correctly) learns to disregard both features, the Entropy-PFI should be zero.

\begin{figure}[ht]
    \centering
    \includegraphics[width=\textwidth]{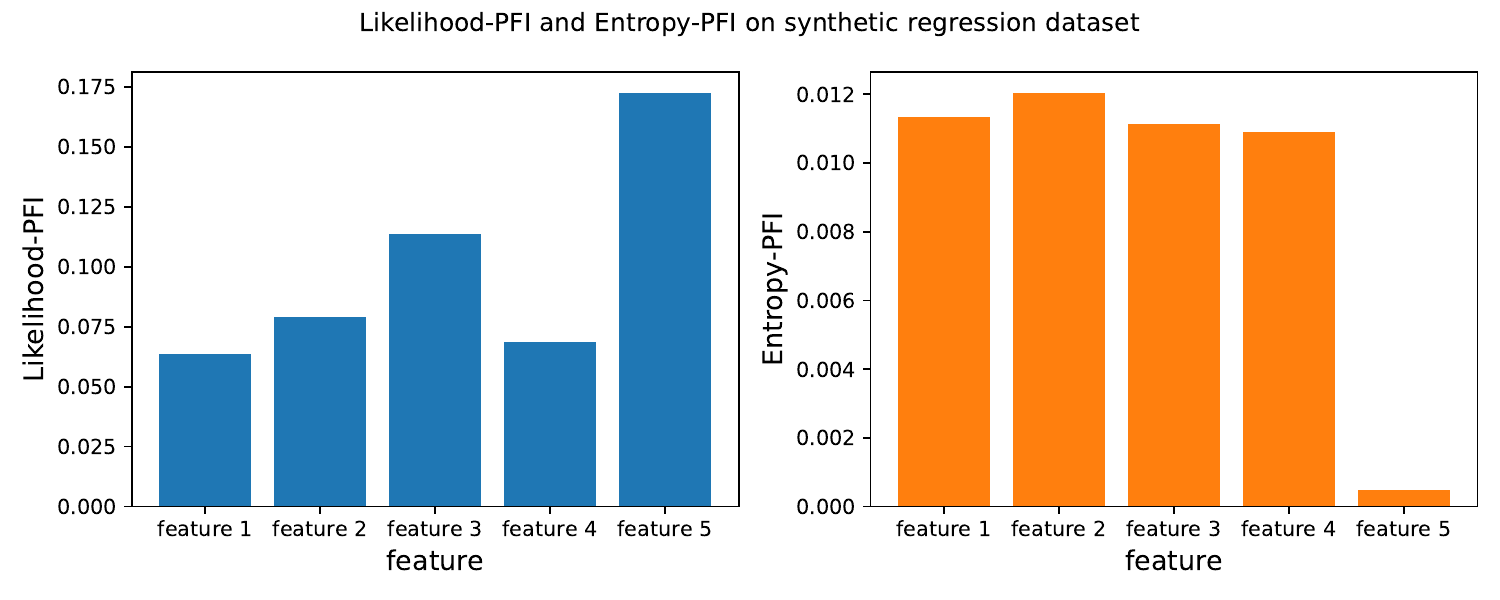}
    \caption{Likelihood-PFI and Entropy-PFI for features in a synthetic dataset using a Gaussian process model. Since features 1-4 share information with each other, their Likelihood-PFI is reduced relative to the independent feature 5. In contrast, their shared information means that they have higher Entropy-PFI, where feature 5's Entropy-PFI is negligible.}
    \label{fig:regression_pfi_toy}
\end{figure}

\paragraph{Regression example on synthetic data} A synthetic dataset is simulated from the regression model
\begin{align*}
    Y = X_1 + X_2 + 0.9 X_3^2 + X_4 + X_5 + \varepsilon,
\end{align*}
where $Y$ is a random variable dependent on feature variables $X_1, \ldots, X_5$.
We sample the features from the following Gaussian distributions:
\begin{equation*}
(X_1, X_2),~(X_3, X_4) \sim \mathcal{N}\left(
\begin{pmatrix}
0\\0
\end{pmatrix},
\begin{pmatrix}
1 & 0.8\\
0.8 & 1
\end{pmatrix}
\right),~
X_5\sim\mathcal{N}(0, 1),~
\varepsilon\sim\mathcal{N}(0, 2).
\end{equation*}
Apart from the stated relationships, the features are otherwise all independent of each other. We train a Gaussian process regression model on 500 examples drawn from this distribution and generate an additional 500 test samples to be used for the importance measures. In Figure~\ref{fig:regression_pfi_toy}, we show the Entropy-PFI and Likelihood-PFI, using the predictive distribution from the Gaussian process. \reviewerThree{Note that this predictive distribution includes uncertainty from the predicted noise, as well as the predicted distribution of the mean.}

In Figure~\ref{fig:regression_pfi_toy}, we show the importance of each feature as measured by Likelihood-PFI and Entropy-PFI. For Likelihood-PFI, we observe many of the known properties of PFI under typical loss functions \citep{molnar2022interpretable}: having the same information shared between multiple features (e.g., having a large covariance between features $X_1$ and $X_2$) diminishes their importance to the model relative to features that contain no shared information (e.g., feature $X_5$). 
In contrast, because the features $X_1$ and $X_2$ contain shared information, their Entropy-PFI is high: when the connection between these features is broken by permuting one of their values in the test set, the model is forced to extrapolate, resulting in higher predictive uncertainty. We see similar behaviour in the measures for $X_3$ and $X_4$, despite their contributions to $Y$ being related in a more complex way. \reviewerThree{In both cases, the Entropy-PFI being high is suggestive of the fact that the Likelihood-PFI is likely lower than if the  feature were independent of all others, as some of the information contained in that feature is also available from other sources.} 

Although we see that feature $X_5$ is considered important in determining the negative log-likelihood (i.e., the model uses information from feature $X_5$ in order to make an accurate prediction), it is not considered important in determining the uncertainty (i.e., on average, knowing feature $X_5$ neither increases nor decreases the model's confidence in its prediction); again, this is due to the feature being independent from the others.

\begin{figure}[ht]
    \centering
    \includegraphics[width=\textwidth]{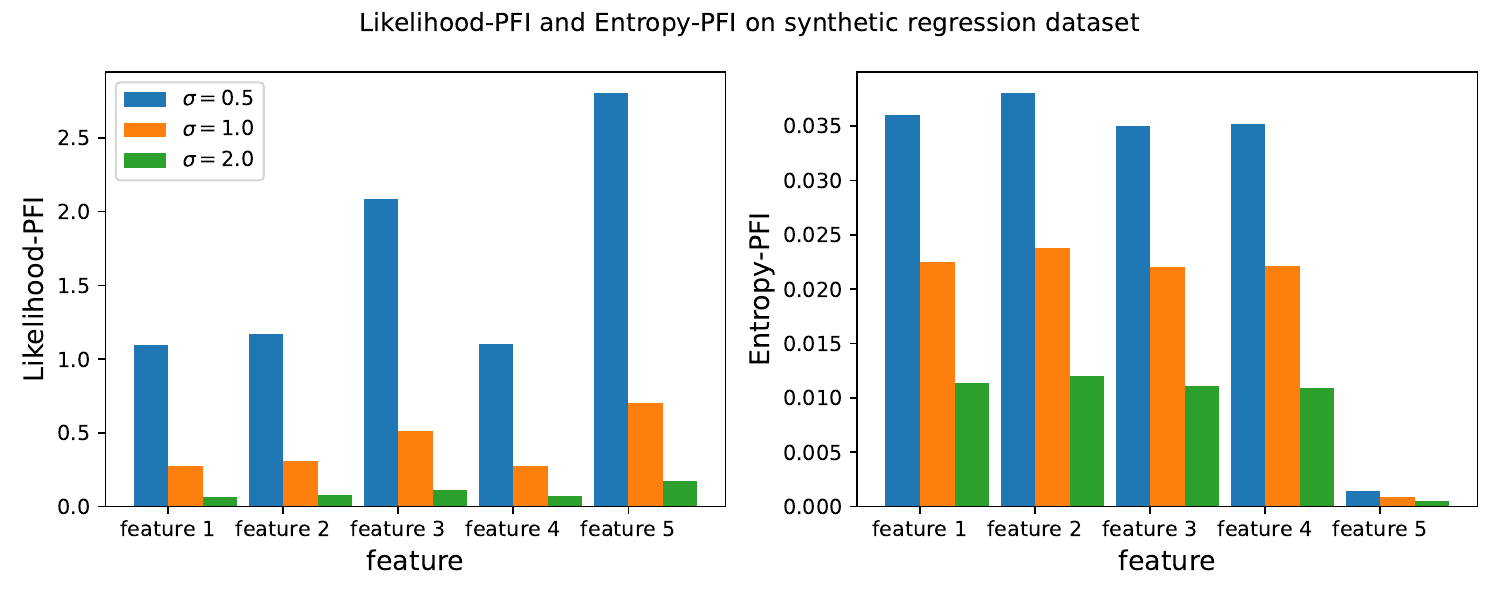}
    \caption{Likelihood-PFI and Entropy-PFI for datasets varying the amount of noise in the target variable. The datasets are the same as in Figure~\ref{fig:regression_pfi_toy}, but with the variance of $\epsilon$ set to the value $\sigma^2$ in each case. We see that when reducing the amount of noise, both the Likelihood-PFI and Entropy-PFI increase.}
    \label{fig:regression_pfi_noise}
\end{figure}

\reviewerThree{Using this dataset, we are also able to demonstrate that the Entropy-PFI is not purely a function of epistemic uncertainty, but that the aleatoric uncertainty of the model also plays a role in determining its value. To this end, we train two more models with similar datasets, but this time changing the amount of noise in the target variable. In particular, we consider the cases where $\varepsilon\sim\mathcal{N}(0, 0.5)$ and $\varepsilon\sim\mathcal{N}(0, 1)$. We plot the results of all three experiments side-by-side in Figure~\ref{fig:regression_pfi_noise}.}

With less noise in the target, we would expect that the predictive distribution of a well-calibrated model to packed more densely around the predictive mean; for this reason, it makes sense that permuting a feature would lead to a prediction which is more harshly punished by the loss, and that the Likelihood-PFI would be larger. Similarly, we observe that the increase in uncertainty when permuting the features is significantly greater.

\section{PDP and ICE for entropy and likelihood}

Entropy-PFI and Likelihood-PFI inherit the property from the original PFI of being global measures of importance. In the previous section, we saw these two measures can complement each other to give useful insights into the global behaviour of a model's predictive distribution. However, there are limitations to this approach. As we saw in Proposition~\ref{prop:independent_pfi}, if a feature's distribution is independent of all others, the Entropy-PFI will be zero even if the uncertainty is greater for some values in the feature's range than others. To alleviate this issue, and to get a more fine-grained explanation of the effects of features on uncertainty, we need an alternate approach.

\subsection{Adapting PDP and ICE for entropy and likelihood}

To tackle the issues discussed above, we examine how we may define partial dependence plots for entropy (Entropy-PDPs). We define these as follows: 
\begin{align}
    \expectedEPDP(x; j) =  \mathbb{E}_{\Xnotj} [\entropy(Y|\Xnotj, X_j=x)],
\end{align}
and we approximate this value using a test set in the following way:
\begin{align}
   \empiricalEPDP(x; j)  =  \frac{1}{n} \sum_{i=1}^n \entropy(Y|\Xnotj=\xnotj^{(i)}, X_j=x).
\end{align}
Entropy-PDP can be reasoned about in the same way as traditional PDP. Given the marginal distribution of $\Xnotj$, the Entropy-PDP at point $\vectorx$ tells us the amount of entropy that we would see in expectation over that distribution if we fixed the $j$th feature's value to $x$.
We similarly define Entropy-ICE plots with
\begin{align}
    \EICE^{(i)}(x; j) = \entropy(Y|\Xnotj=\xnotj^{(i)}, X_j=x).
\end{align}

PDPs and ICE plots tend to be used on the model output itself. Entropy-PDPs and Entropy-ICEs can be seen as extensions of this idea, but using a statistic derived from the model output distribution, as opposed to using the model output directly. 

\reviewerFour{Entropy-PDP shows how sensitive the uncertainty of a model is to changes in the value of the $j$-th feature. Regions where this value is lower can indicate that a feature is strongly informative to the model's output, typically adding confidence to the model's prediction. On the other hand, regions where this value is higher indicate increased uncertainty from the model. This can be due to some feature values being intrinsically linked to higher aleatoric uncertainty, or due to epistemic uncertainty caused by the model being forced to extrapolate for some test examples. These two cases can be distinguished by examining individual ICE curves; in the former case, the ICE curves will show an increase in the region for all test points, while in the latter, we would expect the uncertainty to be lower for curves where the original $x_j$ value falls within that region.}

We now further extend this notion, looking at the likelihood of the true label given the features. In this way we define the PDP for likelihood (Likelihood-PDP) as 
\begin{align*}
   \expectedLPDP(x; j) = -\mathbb{E}_{Y, \Xnotj}[\log q(Y| \Xnotj, X_j=x)].
\end{align*}
Note that this is defined in terms of the negative log-likelihood. The PDP for likelihood is approximated on a test set via averaging:
\begin{align*}
   \empiricalLPDP(x; j) = -\frac{1}{n} \sum_{i=1}^n \log q(y^{(i)}| \Xnotj =\mathbf{x}^{(i)}_{-j}, X_j=x).
\end{align*}

\reviewerThree{While Likelihood-PFI can be thought of as a relatively straightforward adaptation of PFI to an uncertainty-aware setting, Likelihood-PDP is more of a departure from the original PDP, in the sense that while the original PDP measures the value of the function itself, Likelihood-PDP measures the model's performance on a test set and is explicitly target-dependent.} 

Again, we can extend this to ICE curves for the likelihood as follows:

\begin{align*}
   \LICE^{(i)}(x; j) = - \log q(y^{(i)}| \Xnotj =\mathbf{x}^{(i)}_{-j}, X_j=x).
\end{align*}

\reviewerFour{Compared to Entropy-PDP, the interpretation of Likelihood-PDP curves is relatively simple: higher values mean that the model finds it harder to make accurate confident predictions in a region, while lower values mean that it is easier for the model to predict accurately in a region. Likelihood-ICE curves have a similarly simple interpretation: a Likelihood-ICE curve shows how well the model is able to predict the target variable given $\xnotj^{(i)}$ and the given value for $x_j$. Higher regions are regions where the value for $x_j$ misleads the model, decreasing the likelihood of the target, while lower regions indicate that the given value of $x_j$ makes the model more confident and accurate.}

\subsection{A toy example for Entropy-PDP and Entropy-ICE}

\begin{figure}[ht]
    \centering
    \includegraphics[width=\textwidth]{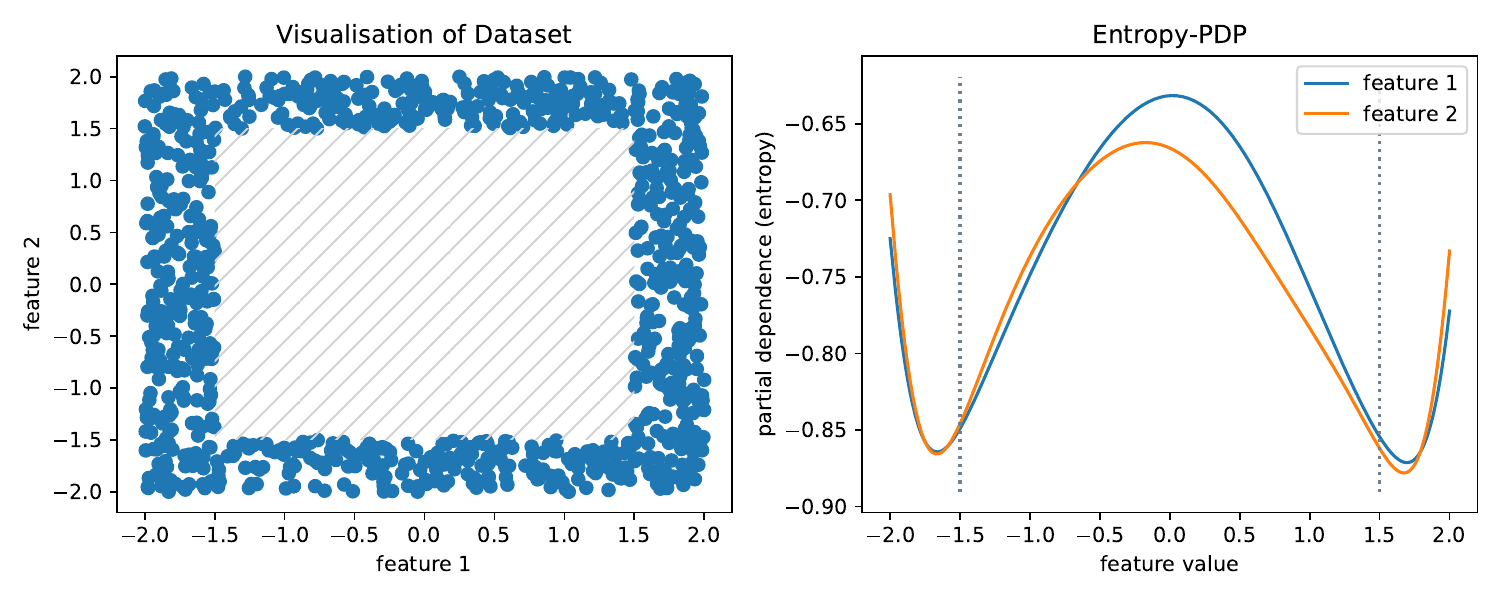}
    \caption{Visualisation of the distribution of synthetic dataset (left) and Entropy-PDP plots for each feature (right). The vertical dotted lines on the Entropy-PDP plot show where the ``interior'' of the distribution begins.}
    \label{fig:synthetic_pdp}
\end{figure}

To offer an interpretation of the Entropy-PDP plots---and in particular to highlight the importance of also plotting Entropy-ICE plots---we consider a synthetic dataset, the distribution of (training) points for which is shown on the left of Figure~\ref{fig:synthetic_pdp}.
We observe that the data is distributed around the border of the feature space, with no examples lying on the interior. This means that an uncertainty-aware model should exhibit high epistemic uncertainty when neither feature takes on an extreme value (i.e., $-1.5 < X_1, X_2 < 1.5$, as shown by the hatched area), and lower uncertainty when either feature takes on a more extreme value \reviewerOne{(i.e., either $|X_1|>1.5$ or $|X_2|>1.5$)}.
The target variable (not shown) is of the form $Y= (X_1 + X_2)^2 + 0.1 \epsilon$, where $\epsilon \sim \mathcal{N}(0, 1)$. Note that by symmetry of the features in the data-generating process, both features should appear equally important to any reasonable model and importance measure; therefore, little insight is to be gained from PFI or similar methods. However, we can still gain an understanding of our model's uncertainty by looking at Entropy-PDP and Entropy-ICE.

In the right-hand plot of Figure~\ref{fig:synthetic_pdp}, we see the Entropy-PDP plot for a Gaussian process model on a test set drawn from the same distribution as the training set. We observe high uncertainty both in the interior values and at very extreme values. We can hypothesise that the uncertainty in the interior is caused by examples where the feature not under consideration is also mid-range, causing the model input to be out-of-distribution and, therefore, for the model to exhibit high epistemic uncertainty. Similarly, at the most extreme values, the uncertainty may be higher, as there are fewer proximal training examples than in the centre of the bands.

This is a case where PDP fails to capture the heterogeneity of the data: we see that uncertainty increases for interior values but have no information about whether this is true for all possible values of the other feature or just some values.

\begin{figure}[ht]
    \centering
    \includegraphics[width=\textwidth]{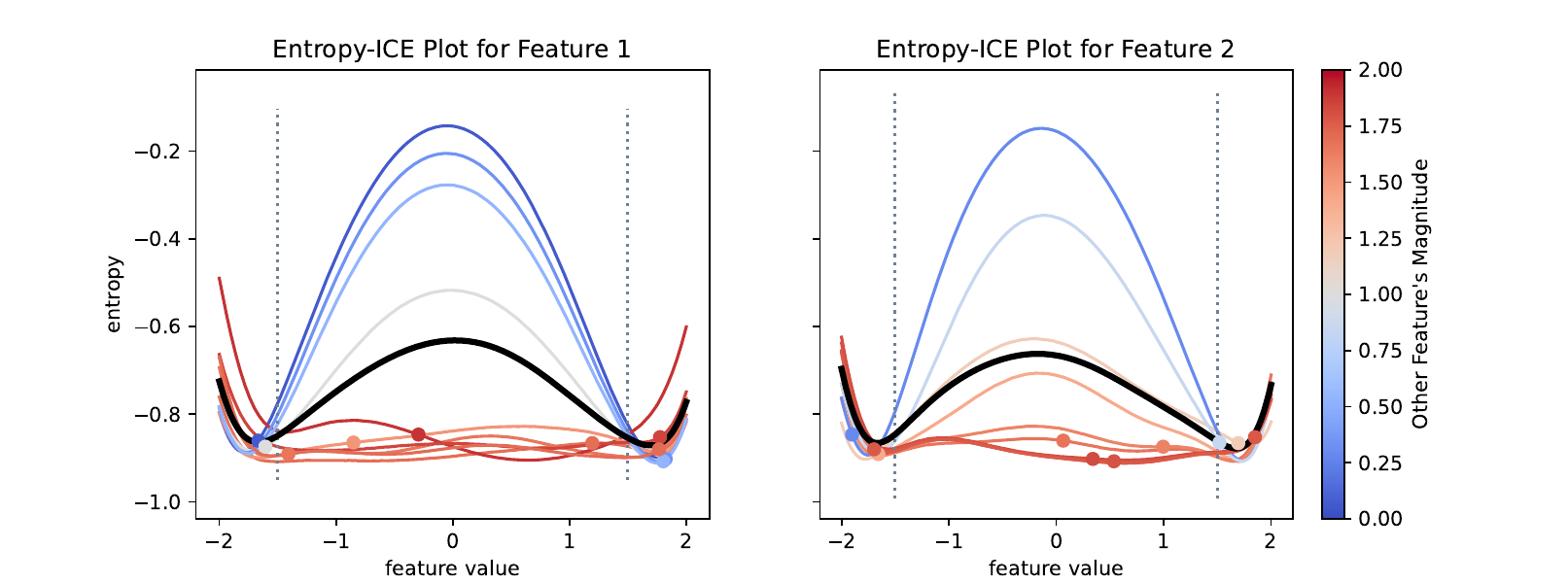}
\caption{Entropy-ICE plots for randomly sampled test points for each feature. The PDP curve is shown in black.
All other lines are ICE plots, with the colour of the line showing the value of the constant feature (i.e., $\xnotj^{(i)}$). The point on each line shows the value of the feature in the original example used to construct the ICE curve.}
    \label{fig:synthetic_ice}
\end{figure}

\reviewerFour{The fact that the Entropy-PDP can be affected by extrapolation in this way is a reason to proceed with caution when analysing these plots. We therefore additionally look at Entropy-ICE plots (Figure~\ref{fig:synthetic_ice}).} In these plots, we can see two distinct behaviours: if we consider the Entropy-ICE plot for the first feature, for examples where feature $X_2$ has small magnitude, we observe higher uncertainty when the feature under consideration ($X_1$) is also small. As hypothesised, this is due to the example being out-of-distribution and, therefore, having high associated uncertainty. On the other hand, when the complementary feature takes a more extreme value, the example generated will be in-distribution and, therefore, have lower associated uncertainty.

\reviewerFour{As an example of how Entropy-ICE curves may also be supplemented with additional information, Figure~\ref{fig:synthetic_ice} shows the values taken by the feature in the examples from which each ICE curve is constructed. This gives us further evidence of extrapolation in Entropy-PDP by demonstrating that none of the original test examples has entropy as high as the peak of the Entropy-PDP.}
We also observe that the behaviour of the ICE curves at extreme values is more uniform: for all test points, the model uniformly becomes more uncertain as the feature value approaches the edge of the distribution, in contrast with the heterogeneous behaviour for central values.

\section{Experiments using real-world datasets}\label{sec:applied}

In this section, we examine how Entropy-PFI and Likelihood-PFI can be used in practice to gain insights into how various probabilistic models make their predictions. We consider a variety of models in both classification and regression settings.

\subsection{Regression example: concrete dataset}\label{sec:concrete}

In this example, we show how the proposed methods can be used to gain insight into the behaviour of models on a real-world regression dataset.
Here, we demonstrate how Likelihood-PFI and Entropy-PFI give complementary explanations for the behaviour of uncertainty-aware regression models. We consider the UCI concrete dataset~\citep{misc_concrete_compressive_strength_165}. We use two uncertainty-aware models: a Gaussian process with a radial basis function (RBF) kernel and a neural network using Monte-Carlo (MC) dropout~\citep{gal2016dropout}. Details of the configurations for both models can be found in Appendix~\ref{sec:hyperparams}. We use 75\% of the dataset for training and the other 25\% for testing.

\begin{figure}[ht]
    \centering
    \includegraphics[width=\textwidth]{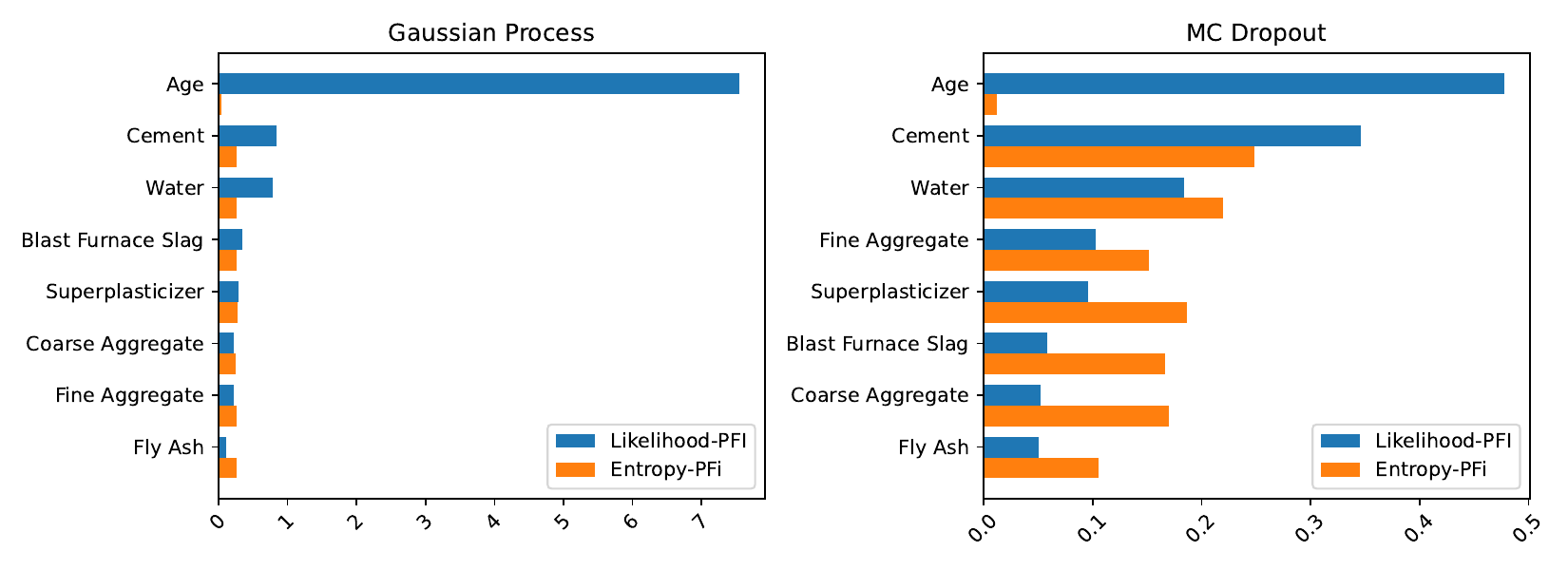}
    \caption{Comparison of Entropy-PFI and Likelihood-PFI for neural networks with Monte-Carlo dropout and Gaussian processes fitted to the UCI concrete dataset.}
    \label{fig:concrete_regression}
\end{figure}

In Figure~\ref{fig:concrete_regression}, we see the relative importance of each feature in terms of both Likelihood-PFI and Entropy-PFI. We observe that although \emph{age} is the most important feature in terms of the likelihood, its (global) effect on the entropy is small. This suggests that age is important in accurately predicting the target variable (i.e., the value of the feature will often have a significant effect on the likelihood), but that it is not strongly related to any of the other features. 

\begin{figure}[ht]
    \centering
    \includegraphics[width=\textwidth]{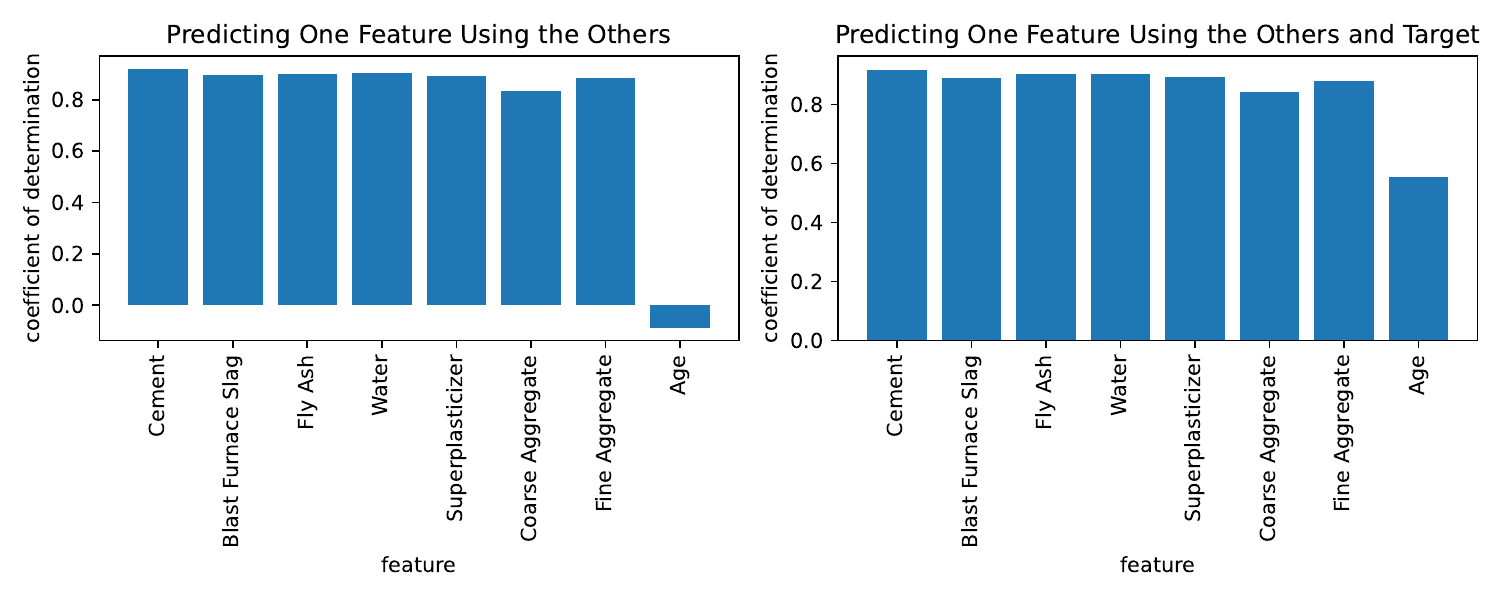}
    \caption{Comparison of coefficients of determination for models predicting one feature's value given the others (UCI concrete dataset).}
    \label{fig:learning_features}
\end{figure}

We can verify this by examining how effectively we can train regression models to learn each feature in the dataset given the others. In Figure~\ref{fig:learning_features}, the coefficient of determination when a random forest regression model model is trained to predict one feature given that we observe all the other features (on the right, this includes the target). We see that indeed \emph{age} appears to be independent of the other features: knowing all the other features does not provide reliable information about age. However, the prediction of the \emph{age} variable improves significantly when we have access to the target variable, suggesting that the target and \emph{age} share information.

Note that in concluding that age is independent of the other features, we draw on several observations. The fact that the Likelihood-PFI is high for the feature discounts the possibility that the Entropy-PFI being zero is simply a result of the model discarding the feature and not making use of it in determining the predictive distribution. Furthermore, the fact that the Entropy-PFI is non-zero for other features means that the model is indeed uncertainty-aware, and is not just using the same distribution with shifted mean for each point.

\begin{figure}[ht]
    \centering
    \includegraphics[width=\textwidth]{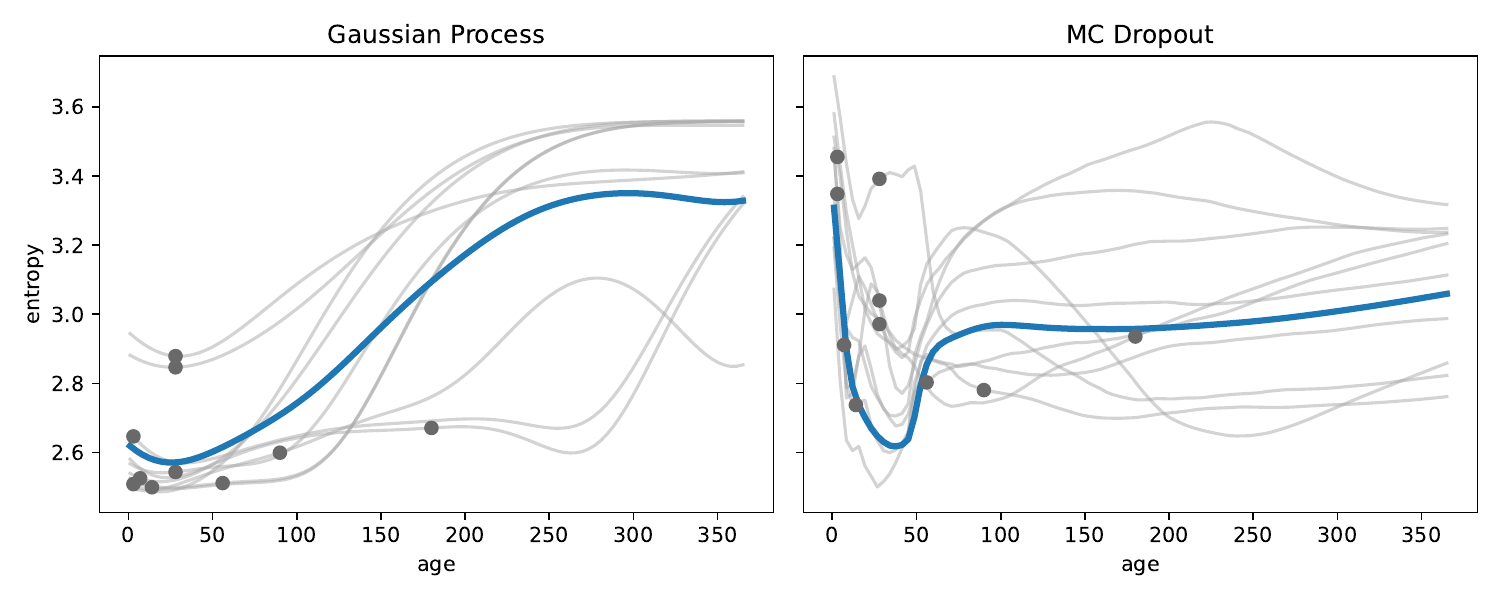}
    \caption{Entropy-ICE and Entropy-PDP plots for \emph{age} feature (UCI concrete dataset). The thicker blue curves are the PDP curves, with ICE curves for examples in the test set shown in grey. The original values for the feature for each ICE curve are shown as grey dots lying on each grey curve.}
    \label{fig:concretePDP}
\end{figure}

However, as previously noted, just because the Entropy-PFI is small/zero, it does not mean that the entropy is not affected \emph{locally} by the specific value of the feature.
To better understand this, in Figure~\ref{fig:concretePDP} we plot the Entropy-PDP curve for \emph{age} for both models, along with a few randomly chosen ICE curves. This figure highlights the fact that Entropy-PFI is a global property: despite Entropy-PFI being near zero, we see that entropy varies not only as we change the feature value (shown by how the PDP curve changes as the value for \emph{age} does), but is also affected by the values that other features take (shown by the variation in characteristics of the ICE curves).

\subsection{Classification example: diabetes dataset}

To demonstrate the utility of our approach for uncertainty-aware models in a classification setting, we examine the importance of various features for models trained on the UCI diabetes dataset \cite{smith1988using}. In particular, we examine two models: calibrated random forests and deep neural networks with weight uncertainty, also known as Bayes by backprop (BBB)~\citep{blundell2015weight}. Further details of the configurations of both models are given in Appendix~\ref{sec:hyperparams}. The UCI diabetes dataset was also used by~\cite{breiman2001random} as one of the first applications of PFI to explain model behaviour. For this example, we first review the findings of~\cite{breiman2001random}, before examining what additional insights could be gleaned from our new approach.

In~\cite{breiman2001random}, it is observed that the second feature (\emph{plas}) is the most important, followed by \emph{age} (feature 8) and \emph{mass} (feature 6). Through additional experiments, Breiman have also shown that while feature 8 contains useful information about the target label, the predictive information that it contains is redundant with the information contained in feature 2; hence, training a model with or without this feature has little effect on the model's predictive power.

\begin{figure}[ht]
    \centering
    \includegraphics[width=\textwidth]{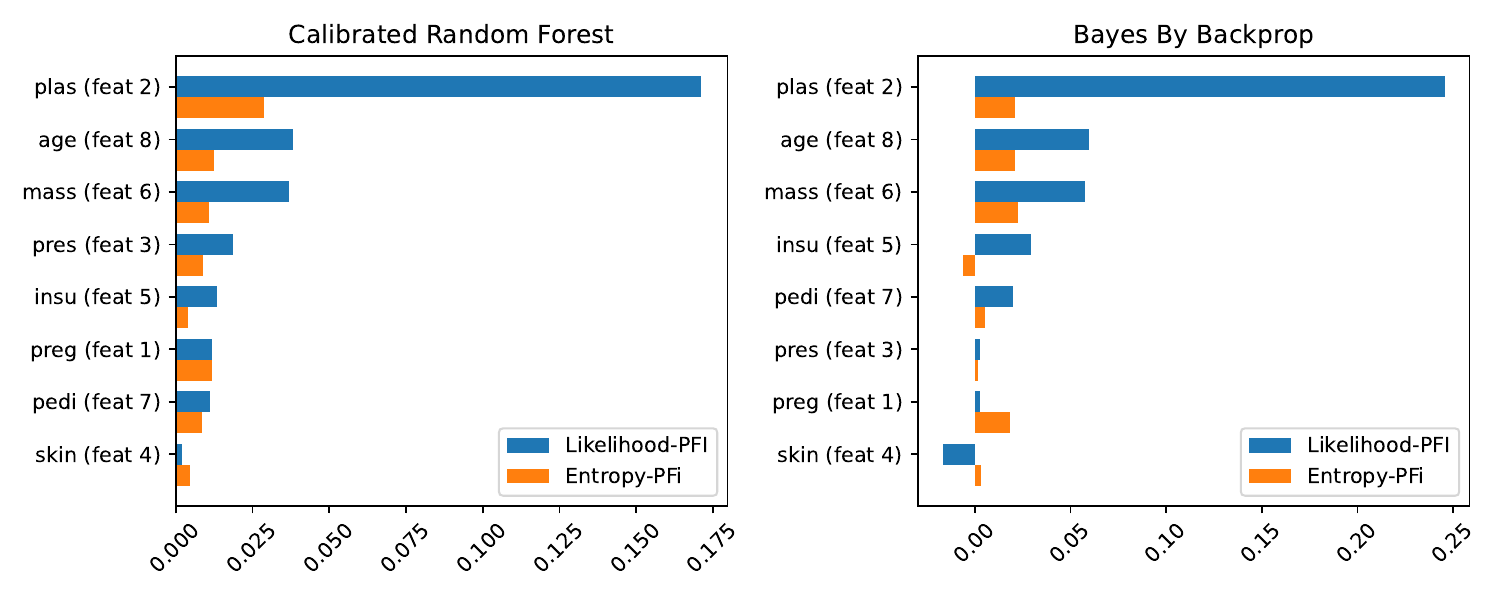}
    \caption{Entropy-PFI and Likelihood-PFI values for calibrated random forests and Bayes by backprop neural networks fitted to the UCI diabetes dataset. Ordered by Likelihood-PFI.}
    \label{fig:diabetes_pfi}
\end{figure}

Where~\cite{breiman2001random} measured the percentage increase in classification accuracy, we measure the difference in likelihood as defined in Equations~\ref{eq:emp_pfi_like_def} and~\ref{eq:exp_pfi_like_def}. Doing so, we observe in Figure~\ref{fig:diabetes_pfi} the same phenomena occur for the likelihood in our random forest model as occurred for classification accuracy in Breiman's: \emph{plas} (feature 2) is the most important, with \emph{age} (feature 8) and \emph{mass} (feature 6) also having significant effects on the model's predictive power. We observe similar results for an MLP using BBB.

\begin{figure}[ht]
    \centering
    \includegraphics[width=\textwidth]{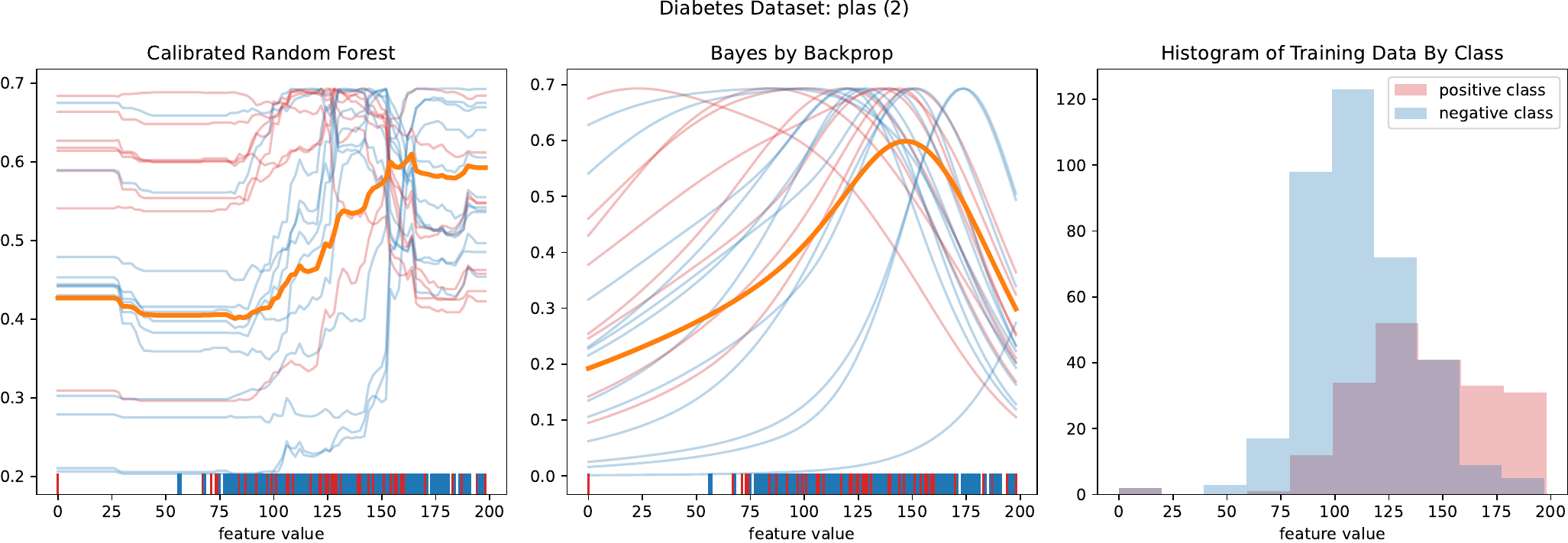}
    \caption{Entropy-PDP (orange) and Entropy-ICE curves  for the UCI diabetes dataset for \emph{plas} (feature 2). Left: curves for a calibrated random forest. Middle: curves for Bayes by backprop. Right: distribution of the feature value in the training set for both classes. In the left and middle figures, the bars along the bottom show values of the feature for examples in the test set of the positive (red) and negative (blue) classes. The red and blue lines are ICE curves for test examples of the two classes, and the orange curves are the Entropy-PDP curves.}
    \label{fig:diabetes_pdp_plas}
\end{figure}

In Figure~\ref{fig:diabetes_pdp_plas}, we show the Entropy-PDP and Entropy-ICE plots for \emph{plas} (feature 2). We see that for lower values of the feature, there is a relatively low entropy for both models. We also observe a significant increase in uncertainty for higher values, peaking at about 150. Examining the distribution of the feature in the training set, separated by class, we see that at extreme values, the feature is strongly informative of class label, but at values in the region 100-170, both classes are present with high frequency, leading the feature to be less informative and therefore rendering models less confident in their predictions based on information from this feature for values in this range.

\begin{figure}[ht]
    \centering
    \includegraphics[width=\textwidth]{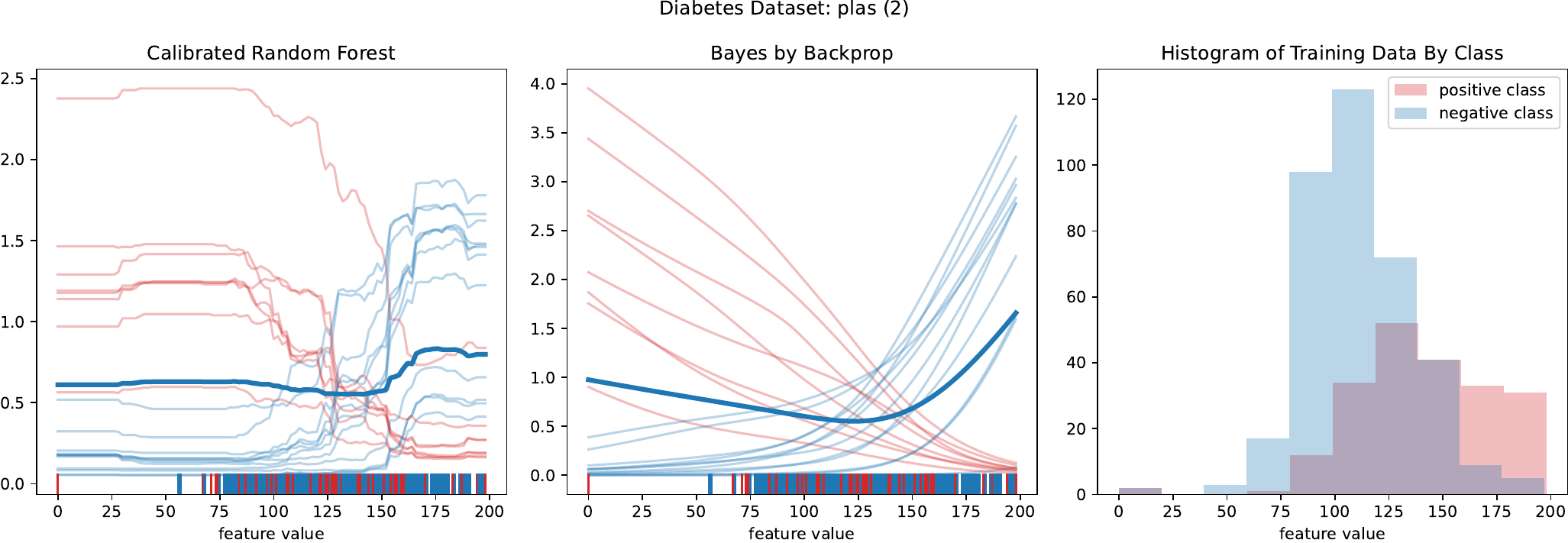}
    \caption{Likelihood-PDP (blue, bold) and Likelihood-ICE curves  for the UCI diabetes dataset for \emph{plas} (feature 2). Left: curves for calibrated random forest. Middle: curves for Bayes by backprop. Right: distribution of feature value in training set for both classes.}
    \label{fig:diabetes_pdp_plas_like}
\end{figure}

We can also see the effect of this change in the confidence level of the models on the negative log-likelihood in Figure~\ref{fig:diabetes_pdp_plas_like}. For examples in the positive class, the model becomes more confident in its correct prediction (given the other features) as the \emph{plas} value increases, leading to a decrease in the loss for those examples. For the negative class, the opposite is true: The model becomes more confident in its prediction as the feature value decreases. Again, we can interpret this in terms of the confidence increasing as the feature of interest adds evidence to support the conclusion inferred from the other features.

\begin{figure}[ht]
    \centering
    \includegraphics[width=\textwidth]{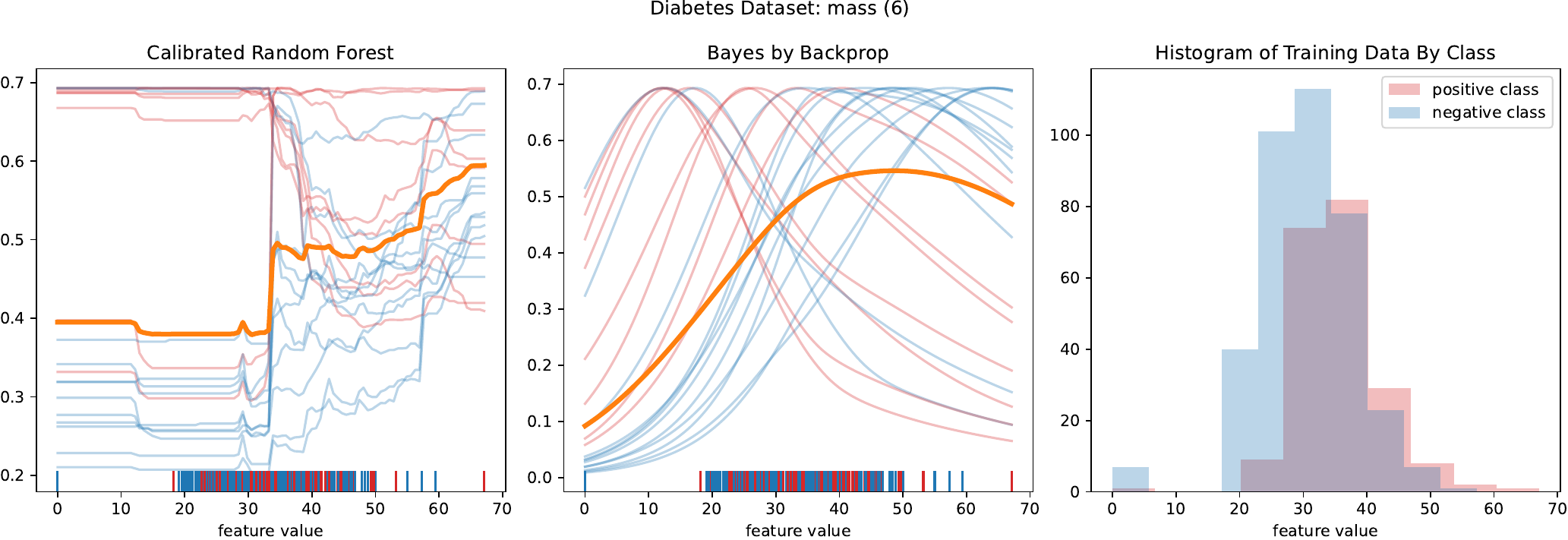}
    \caption{Entropy-PDP (orange) and Entropy-ICE curves for the UCI diabetes dataset for \emph{mass} (feature 6). Left: curves for a calibrated random forest. Middle: curves for Bayes by backprop. Right: distribution of feature value in training set for both positive and negative classes.}
    \label{fig:diabetes_pdp_mass}
\end{figure}

We observe a similar phenomenon for \emph{mass} (feature 6) in Figure~\ref{fig:diabetes_pdp_mass}. In particular, for members of the positive class, the ICE curves generally show a greater amount of uncertainty for low feature values. 
This may be due to the training set not containing examples from the positive class where this value is low and therefore the examples constructed for the ICE curve are out-of-distribution and exhibit high epistemic uncertainty. Note that this is not picked up in the Entropy-PDP, and can only be observed using Entropy-ICE.

\begin{figure}[ht]
    \centering
    \includegraphics[width=\textwidth]{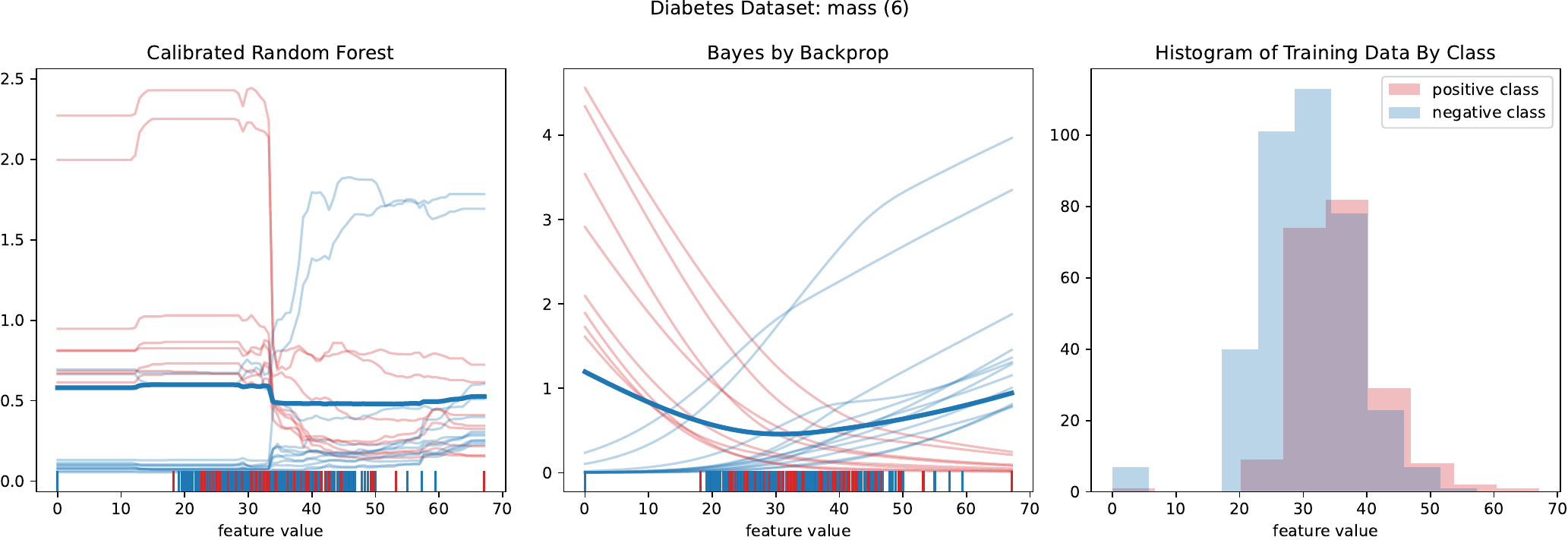}
    \caption{Likelihood-PDP (blue, bold) and Likelihood-ICE curves for the UCI diabetes dataset for \emph{mass} (feature 6). Left: curves for a calibrated random forest. Middle: curves for Bayes by backprop. Right: distribution of feature value in training set for both positive and negative classes.}
    \label{fig:diabetes_pdp_mass_like}
\end{figure}

In Figure~\ref{fig:diabetes_pdp_mass_like}, we again see a difference in behaviour of the Likelihood-ICE curves for the different classes, with the loss being lower for the positive class at higher values for the feature, and for the negative class at lower values.

\newpage
\section{Conclusions}~\label{sec:conclusion}

In this paper, we have proposed modifications of PFI, PDP and ICE that can be used to gain insights into the importance of features in uncertainty-aware models, both in terms of likelihood and uncertainty (as measured by the entropy of the predictive distribution).

Permutation feature importance, amongst other methods, has come under criticism for their shortcomings in forcing the model to extrapolate to unexplored regions in developing explanations. Although the suggested solution is to avoid PFI in favour of methods that explicitly address this issue, the simplicity of PFI and related approaches mean that they nonetheless remain popular. Using Entropy-PFI can mitigate some of these issues, providing additional information about the level of uncertainty in a model that is attributable to a given feature. In particular, Entropy-PFI can be used to identify when a feature is likely to be independent of other informative features, and therefore its feature importance can be trusted.

We note that Entropy-PFI does not completely mitigate the issues raised in~\citep{hooker2021unrestricted}, and careful interpretation and understanding of the strengths and weaknesses of each method are required. However, given the prevalence of permutation and extrapolation-based importance methods, even in light of recent criticism, having these additional tools serves to mitigate some of the shortcomings of these approaches.

\reviewerThree{Returning to the medical example we gave in Section~\ref{sec:introduction}, we now have the tools to analyse our hypothetical model. By examining PDP and ICE curves, we can determine if the given feature is the source of the uncertainty based on how the uncertainty changes as we vary its value. Additionally, we can use Entropy-PFI to determine if the uncertainty is due to disagreements between features.}

Throughout, when demonstrating entropy-based feature importance methods, we found it useful in building intuition to make appeals to the readers' understanding of aleatoric and epistemic uncertainty. 
However, we did not attempt to separate them. The utility of explaining these two sources of uncertainty, as well as the best methodology for doing so, remain questions for future research.

Similarly, there are many other methods that could be used for explainable uncertainty, such as LIME~\citep{ribeiro2016should}, and adaptations of those used in this paper (such as showing derivatives using ICE rather than the original values).

\backmatter


\bmhead{Funding}
The authors acknowledge financial support from The University of Manchester's Centre for Digital Trust and Society and Simon Industrial Fund. 


\bmhead{Code availability}
Our code for running the experiments of this paper is available at \url{https://github.com/EchoStatements/entropy-based-importance-methods}.







\begin{appendices}

\section{A note on notation}\label{app:regarding_notation}

In this appendix, we further clarify the notation used in defining expectations in Likelihood-PFI and Entropy-PFI. In conventional PFI with loss function $\mathcal{C}$, and model output $f(X) = f(\Xj, \Xnotj)$, we would write
\begin{align*}
    PFI(j) = \mathbb{E}_{Y, X,\Xcopy}[\mathcal{C}(Y, f(\Xcopy_j, \Xnotj)) - \mathcal{C}(Y, f(X))].
\end{align*}
However, for the models we are interested in, the output of the model is not a scalar (or vector)-valued function, but a density $q(Y|X)$. For a given test point, we can write $q(Y=y|X=x)$ as the probability assigned to the label $y$ given the features $x$. Additionally, we can define the Likelihood-PFI for a single point $(\vectorx, y)$ as 
\begin{align*}
    \expectedEPFI(\vectorx; j) = -\int  \log q(Y=y | X_j=\widetilde{x}_j, \Xnotj=\xnotj)\, P_{\Xcopy_j}(\widetilde{x}_j) \, d\widetilde{x}_j + \log q(Y=y|X=\vectorx) , 
\end{align*}
remembering that $X=(X_j, \Xnotj)$ and that $\Xcopy_j$ is an independent random variable whose distribution is the same as the marginal distribution of $X_j$.

This quantity depends on the particular test point $(\vectorx, y)$ and on the distribution $P_{\Xcopy_j}$. Now say that we want to find the expectation of this quantity over the true data distribution, whose density is denoted as $P_{X,Y}$. For the second term, we may simply write
\begin{align*}
    \mathbb{E}_{X, Y} [\log q(Y|X)] = \int \log q(Y=y|X=\vectorx) P_{X, Y}(\vectorx,y) \,d\vectorx.
\end{align*}
The notation is natural because the density function used is the true density of $X$. 
However, note that in the above definition for PFI at a single point, we have integrated out $X_j$ already, but rather than using its true distribution, we have integrated it out acting as though its distribution is that of $\widetilde{X}_j$. This is what we reflect in our notation with $X_j=\widetilde{X}_j$, which serves as shorthand for integrals of the following form
\begin{align*}
    & \mathbb{E}_{\Xnotj, \Xcopy_j, Y}[-\log q(Y|X_{-j}, X_j=\widetilde{X}_j)] \\
    &\defeq -\int\int \int  \log q(Y=y | X_j=\widetilde{x}_j, \Xnotj=\xnotj)\,  P_{\widetilde{X}_j, \Xnotj, Y}(\widetilde{x}_j, x_{-j},y) \,d\widetilde{x}_j\, d\vectorx_{-j}\,dy \\
    &=-\int\int \int  \log q(Y=y | X_j=\widetilde{x}_j, \Xnotj=\xnotj)\, P_{\Xcopy_j}(\widetilde{x}_j) \,  P_{\Xnotj, Y}(x_{-j},y) \,d\widetilde{x}_j\, d\vectorx_{-j}\,dy .
\end{align*}

\section{Model and hyperparameter setup}\label{sec:hyperparams}

\reviewerThree{In this appendix, we briefly outline the models used in the experiment throughout the paper. In all cases, we use the predictive distribution produced by the model. For classification models, these are the probability estimates assigned to each class, and for regression models, it is the distribution over real values that the model predicts given the features (in our cases, it is a Gaussian distribution).}

\subsection{Gaussian processes}

We use an exact Gaussian process (using GPyTorch~\cite{gardner2018gpytorch}) with a constant mean function and covariance kernel of an RBF kernel composed with a scale kernel.
For hyperparameter optimisation, we use Adam on the hyperparameters with a learning rate of 0.1. For the experiment in Section~\ref{sec:pfi} we train for 1000 epochs, for the experiment in Section~\ref{sec:applied}, we train for 200 epochs.

\subsection{Calibrated random forests}

We use the scikit-learn~\citep{scikit-learn} implementation of random forests with 500 trees with maximum depth 8. 80\% of training data is used to train the forest itself, with the remaining 20\% used for calibration based on scikit-learn's \texttt{CalibratedClassifierCV} with the ``sigmoid'' method.

\subsection{MLP with Bayes by backprop}

We use an MLP with two hidden layers with 100 units each. Weights are given a Gaussian prior with mean zero and standard deviation $0.5$.
The MLP is trained for 100 epochs with a learning rate of 0.01.

\subsection{MLP with MC dropout}

We use an MLP with two hidden layers with 100 units each and a dropout rate of $0.1$. All features and targets are scaled so that the training set has zero mean and unit variance, and outputs are re-scaled to their original values for model evaluation metrics.
The MLP is trained using SGD with learning rate 0.1 for 5000 epochs with a batch size of 200.
Furthermore, the reserved 10\% of the training data is used to tune the precision parameter $\tau$ by gradient descent.

\subsection{Random forest regression model}

The random forest regression model used in Section~\ref{sec:concrete} is a random forest of 100 trees with no maximum depth. The scikit-learn \texttt{RandomForestRegressor} model is used with default arguments.




\end{appendices}



\input{main.bbl}
\end{document}

%% file: definitions.tex
\definecolor{Fuschia}{HTML}{8C368C}
\definecolor{BrickRed}{HTML}{B6321C}
\definecolor{DarkGreen}{HTML}{308610}

\newcommand{\reviewerOne}[1]{\textcolor{black}{#1}}

\newcommand{\reviewerThree}[1]{\textcolor{black}{#1}}
\newcommand{\reviewerFour}[1]{\textcolor{black}{#1}}
\DeclarePairedDelimiterX{\KLx}[2]{(}{)}{%
#1\,\delimsize\|\,#2%
}
\newcommand{\PFI}{\textnormal{PFI}}
\newcommand{\cPFI}{\textnormal{CPFI}}

\newcommand{\entropy}{{\mathcal{H}_q}}
\newcommand{\entropysub}{\mathcal{H}}
\newcommand{\likelihood}{\mathcal{L}}
\newcommand{\loss}{\mathcal{C}}

\newcommand{\expectedPFI}{\PFI_\loss}
\newcommand{\expectedLPFI}{\PFI_\likelihood}
\newcommand{\expectedEPFI}{\PFI_\entropysub}

\newcommand{\expectedcPFI}{\cPFI_\loss}

\newcommand{\expectedcEPFI}{\cPFI_\entropysub}

\newcommand{\empiricalPFI}{\widehat{\PFI}_\loss}
\newcommand{\empiricalEPFI}{\widehat{\PFI}_\entropysub}
\newcommand{\empiricalLPFI}{\widehat{\PFI}_\likelihood}

\newcommand{\expectedPDP}{\textnormal{PDP}}
\newcommand{\empiricalPDP}{\widehat{\textnormal{PDP}}}

\newcommand{\expectedEPDP}{\textnormal{PDP}_\entropysub}
\newcommand{\empiricalEPDP}{\widehat{\textnormal{PDP}}_\entropysub}

\newcommand{\expectedLPDP}{\textnormal{PDP}_\likelihood}
\newcommand{\empiricalLPDP}{\widehat{\textnormal{PDP}}_\likelihood}

\newcommand{\ICE}{\textnormal{ICE}}
\newcommand{\EICE}{{\textnormal{ICE}_\entropysub}}
\newcommand{\LICE}{{\textnormal{ICE}_\likelihood}}

\newcommand{\Xcopy}{\widetilde{X}}
\newcommand{\xcopy}{\widetilde{x}}

{}

\newcommand\defeq{\mathrel{\overset{\makebox[0pt]{\mbox{\normalfont\tiny\sffamily def}}}{=}}}

\newcommand{\Xjcond}{\widetilde{X}^c_j}

\newcommand{\Xj}{X_j}
\newcommand{\Xnotj}{{X_{-j}}}

\newcommand{\xnotj}{\vectorx_{-j}}

\newcommand{\vectorx}{\mathbf{x}}

%% file: main.bbl
\begin{thebibliography}{37}
\providecommand{\natexlab}[1]{#1}
\providecommand{\url}[1]{{#1}}
\providecommand{\urlprefix}{URL }
\providecommand{\doi}[1]{\url{https://doi.org/#1}}
\providecommand{\eprint}[2][]{\url{#2}}
 \bibcommenthead

\bibitem[{Antoran et~al(2021)Antoran, Bhatt, Adel, Weller, and
  Hern{\'a}ndez-Lobato}]{antoran2021getting}
Antoran J, Bhatt U, Adel T, et~al (2021) Getting a {CLUE}: a method for
  explaining uncertainty estimates. In: International Conference on Learning
  Representations

\bibitem[{Blundell et~al(2015)Blundell, Cornebise, Kavukcuoglu, and
  Wierstra}]{blundell2015weight}
Blundell C, Cornebise J, Kavukcuoglu K, et~al (2015) Weight uncertainty in
  neural networks. In: International Conference on Machine Learning

\bibitem[{Breiman(2001)}]{breiman2001random}
Breiman L (2001) Random forests. Machine learning 45:5--32.
  \doi{https://doi.org/10.1023/A:1010933404324}

\bibitem[{Casalicchio et~al(2018)Casalicchio, Molnar, and
  Bischl}]{casalicchio2019visualizing}
Casalicchio G, Molnar C, Bischl B (2018) Visualizing the feature importance for
  black box models. In: Machine Learning and Knowledge Discovery in Databases:
  European Conference, ECML PKDD, Springer, pp 655--670,
  \doi{https://doi.org/10.1007/978-3-030-10925-7_40}

\bibitem[{Chai(2018)}]{Chai2018UncertaintyEI}
Chai LR (2018) Uncertainty estimation in {B}ayesian neural networks and links
  to interpretability. Master's thesis, University of Cambridge

\bibitem[{Chau et~al(2024)Chau, Muandet, and Sejdinovic}]{chau2024explaining}
Chau SL, Muandet K, Sejdinovic D (2024) Explaining the uncertain: Stochastic
  shapley values for gaussian process models. Advances in Neural Information
  Processing Systems 36

\bibitem[{Chen et~al(2023)Chen, Covert, Lundberg, and Lee}]{chen2023algorithms}
Chen H, Covert IC, Lundberg SM, et~al (2023) Algorithms to estimate {S}hapley
  value feature attributions. Nature Machine Intelligence pp 1--12.
  \doi{https://doi.org/10.1038/s42256-023-00657-x}

\bibitem[{Covert et~al(2021)Covert, Lundberg, and Lee}]{covert2021explaining}
Covert IC, Lundberg S, Lee SI (2021) Explaining by removing: a unified
  framework for model explanation. The Journal of Machine Learning Research
  22(1):9477--9566

\bibitem[{Depeweg et~al(2017)Depeweg, Hern{\'a}ndez-Lobato, Udluft, and
  Runkler}]{depeweg2017sensitivity}
Depeweg S, Hern{\'a}ndez-Lobato JM, Udluft S, et~al (2017) Sensitivity analysis
  for predictive uncertainty in {B}ayesian neural networks. arXiv preprint
  arXiv:171203605

\bibitem[{Depeweg et~al(2018)Depeweg, Hernandez-Lobato, Doshi-Velez, and
  Udluft}]{depeweg2018decomposition}
Depeweg S, Hernandez-Lobato JM, Doshi-Velez F, et~al (2018) Decomposition of
  uncertainty in {B}ayesian deep learning for efficient and risk-sensitive
  learning. In: International Conference on Machine Learning

\bibitem[{Friedman(2001)}]{friedman2001greedy}
Friedman JH (2001) Greedy function approximation: a gradient boosting machine.
  Annals of Statistics pp 1189--1232.
  \doi{https://doi.org/10.1214/aos/1013203451}

\bibitem[{Gal and Ghahramani(2016)}]{gal2016dropout}
Gal Y, Ghahramani Z (2016) Dropout as a {B}ayesian approximation: representing
  model uncertainty in deep learning. In: International Conference on Machine
  Learning, pp 1050--1059

\bibitem[{Gardner et~al(2018)Gardner, Pleiss, Bindel, Weinberger, and
  Wilson}]{gardner2018gpytorch}
Gardner JR, Pleiss G, Bindel D, et~al (2018) G{P}y{T}orch: Blackbox
  matrix-matrix {G}aussian process inference with {GPU} acceleration. In:
  Advances in Neural Information Processing Systems

\bibitem[{Goldstein et~al(2015)Goldstein, Kapelner, Bleich, and
  Pitkin}]{goldstein2015peeking}
Goldstein A, Kapelner A, Bleich J, et~al (2015) Peeking inside the black box:
  visualizing statistical learning with plots of individual conditional
  expectation. journal of Computational and Graphical Statistics 24(1):44--65.
  \doi{https://doi.org/10.1080/10618600.2014.907095}

\bibitem[{Guo et~al(2017)Guo, Pleiss, Sun, and Weinberger}]{guo2017calibration}
Guo C, Pleiss G, Sun Y, et~al (2017) On calibration of modern neural networks.
  In: International Conference on Machine Learning, pp 1321--1330

\bibitem[{Hooker et~al(2021)Hooker, Mentch, and Zhou}]{hooker2021unrestricted}
Hooker G, Mentch L, Zhou S (2021) Unrestricted permutation forces
  extrapolation: variable importance requires at least one more model, or there
  is no free variable importance. Statistics and Computing 31:1--16.
  \doi{https://doi.org/10.1007/s11222-021-10057-z}

\bibitem[{Kelly et~al(2020)Kelly, Sachan, Ni, Almaghrabi, Allmendinger, and
  Chen}]{kelly2020explainable}
Kelly L, Sachan S, Ni L, et~al (2020) Explainable artificial intelligence for
  digital forensics: opportunities, challenges and a drug testing case study.
  Digital Forensic Science \doi{https://doi.org/10.5772/intechopen.93310}

\bibitem[{Liu et~al(2019)Liu, Paisley, Kioumourtzoglou, and
  Coull}]{liu2019accurate}
Liu J, Paisley J, Kioumourtzoglou MA, et~al (2019) Accurate uncertainty
  estimation and decomposition in ensemble learning. In: Advances in Neural
  Information Processing Systems

\bibitem[{Lundberg and Lee(2017)}]{lundberg2017shap}
Lundberg SM, Lee SI (2017) A unified approach to interpreting model
  predictions. In: Advances in Neural Information Processing Systems

\bibitem[{Lundberg et~al(2020)Lundberg, Erion, Chen, DeGrave, Prutkin, Nair,
  Katz, Himmelfarb, Bansal, and Lee}]{lundberg2020treeshort}
Lundberg SM, Erion G, Chen H, et~al (2020) From local explanations to global
  understanding with explainable {AI} for trees. Nature Machine Intelligence
  2(1):56--67. \doi{https://doi.org/10.1038/s42256-019-0138-9}

\bibitem[{Mease and Wyner(2008)}]{mease2008evidence}
Mease D, Wyner A (2008) Evidence contrary to the statistical view of boosting.
  Journal of Machine Learning Research 9(2)

\bibitem[{Molnar(2022)}]{molnar2022interpretable}
Molnar C (2022) Interpretable Machine Learning, 2nd edn. Independently
  Published

\bibitem[{Molnar et~al(2023)Molnar, K{\"o}nig, Bischl, and
  Casalicchio}]{molnar2023model}
Molnar C, K{\"o}nig G, Bischl B, et~al (2023) Model-agnostic feature importance
  and effects with dependent features: a conditional subgroup approach. Data
  Mining and Knowledge Discovery pp 1--39.
  \doi{https://doi.org/10.1007/s10618-022-00901-9}

\bibitem[{Moosbauer et~al(2021)Moosbauer, Herbinger, Casalicchio, Lindauer, and
  Bischl}]{moosbauer2021explaining}
Moosbauer J, Herbinger J, Casalicchio G, et~al (2021) Explaining hyperparameter
  optimization via partial dependence plots. In: Advances in Neural Information
  Processing Systems

\bibitem[{Mukhoti et~al(2023)Mukhoti, Kirsch, van Amersfoort, Torr, and
  Gal}]{mukhoti2023deep}
Mukhoti J, Kirsch A, van Amersfoort J, et~al (2023) Deep deterministic
  uncertainty: a new simple baseline. In: Proceedings of the IEEE/CVF
  Conference on Computer Vision and Pattern Recognition,
  \doi{https://dx.doi.org/10.1109/CVPR52729.2023.02336}

\bibitem[{Neal(2012)}]{neal2012bayesian}
Neal RM (2012) Bayesian learning for neural networks, vol 118. Springer Science
  \& Business Media, \doi{https://doi.org/10.1007/978-1-4612-0745-0}

\bibitem[{Pedregosa et~al(2011)Pedregosa, Varoquaux, Gramfort, Michel, Thirion,
  Grisel, Blondel, Prettenhofer, Weiss, Dubourg, Vanderplas, Passos,
  Cournapeau, Brucher, Perrot, and Duchesnay}]{scikit-learn}
Pedregosa F, Varoquaux G, Gramfort A, et~al (2011) Scikit-learn: machine
  learning in {P}ython. Journal of Machine Learning Research 12:2825--2830

\bibitem[{Ribeiro et~al(2016)Ribeiro, Singh, and Guestrin}]{ribeiro2016should}
Ribeiro MT, Singh S, Guestrin C (2016) `{W}hy should {I} trust you?'
  {E}xplaining the predictions of any classifier. In: ACM SIGKDD International
  Conference on Knowledge Discovery and Data Mining, pp 1135--1144,
  \doi{https://doi.org/10.1145/2939672.2939778}

\bibitem[{Shaker and H{\"u}llermeier(2020)}]{shaker2020aleatoric}
Shaker MH, H{\"u}llermeier E (2020) Aleatoric and epistemic uncertainty with
  random forests. In: Advances in Intelligent Data Analysis XVIII: 18th
  International Symposium on Intelligent Data Analysis,
  \doi{https://doi.org/10.1007/978-3-030-44584-3_35}

\bibitem[{Slack et~al(2020)Slack, Hilgard, Jia, Singh, and
  Lakkaraju}]{slack2020fooling}
Slack D, Hilgard S, Jia E, et~al (2020) Fooling {LIME} and {SHAP}: adversarial
  attacks on post hoc explanation methods. In: AAAI/ACM Conference on AI,
  Ethics, and Society, \doi{https://doi.org/10.1145/3375627.3375830}

\bibitem[{Smith et~al(1988)Smith, Everhart, Dickson, Knowler, and
  Johannes}]{smith1988using}
Smith JW, Everhart JE, Dickson W, et~al (1988) Using the {ADAP} learning
  algorithm to forecast the onset of diabetes mellitus. In: Annual Symposium on
  Computer Application in Medical Care, p 261

\bibitem[{Strobl et~al(2008)Strobl, Boulesteix, Kneib, Augustin, and
  Zeileis}]{strobl2008conditional}
Strobl C, Boulesteix AL, Kneib T, et~al (2008) Conditional variable importance
  for random forests. BMC Bioinformatics 9:1--11.
  \doi{https://doi.org/10.1186/1471-2105-9-307}

\bibitem[{Watson et~al(2023)Watson, O'Hara, Tax, Mudd, and
  Guy}]{watson2023explaining}
Watson DS, O'Hara J, Tax N, et~al (2023) Explaining predictive uncertainty with
  information theoretic {S}hapley values. arXiv preprint arXiv:230605724

\bibitem[{Williams and Rasmussen(2006)}]{williams2006gaussian}
Williams CK, Rasmussen CE (2006) Gaussian processes for machine learning. 3,
  MIT press Cambridge, MA, \doi{https://doi.org/10.7551/mitpress/3206.001.0001}

\bibitem[{Wimmer et~al(2023)Wimmer, Sale, Hofman, Bischl, and
  H{\"u}llermeier}]{wimmer2023quantifying}
Wimmer L, Sale Y, Hofman P, et~al (2023) Quantifying aleatoric and epistemic
  uncertainty in machine learning: are conditional entropy and mutual
  information appropriate measures? In: Uncertainty in Artificial Intelligence

\bibitem[{Yeh(2007)}]{misc_concrete_compressive_strength_165}
Yeh IC (2007) Concrete compressive strength. UCI Machine Learning Repository,
  \doi{https://doi.org/10.24432/C5PK67}

\bibitem[{Zhang et~al(2022)Zhang, Chan, and Mahadevan}]{zhang2022explainable}
Zhang X, Chan FT, Mahadevan S (2022) Explainable machine learning in image
  classification models: an uncertainty quantification perspective.
  Knowledge-Based Systems 243:108418.
  \doi{https://doi.org/10.1016/j.knosys.2022.108418}

\end{thebibliography}
